\documentclass[10pt,twocolumn,letterpaper]{article}
\usepackage[]{cvpr}
\definecolor{cvprblue}{rgb}{0.21,0.49,0.74}
\usepackage[pagebackref,breaklinks,colorlinks,allcolors=cvprblue]{hyperref}

\def\paperID{11960}% *** Enter the Paper ID here
\def\confName{CVPR}
\def\confYear{2025}

\usepackage[utf8]{inputenc} % allow utf-8 input
\usepackage[T1]{fontenc}    % use 8-bit T1 fonts
\usepackage{hyperref}       % hyperlinks
\usepackage{comment}
\usepackage{url}            % simple URL typesetting
\usepackage{graphicx}
\usepackage{caption}
\usepackage{graphicx}
\usepackage{makecell}
\usepackage{booktabs}       % professional-quality tables
\usepackage{amsfonts}       % blackboard math symbols
\usepackage{nicefrac}       % compact symbols for 1/2, etc.
\usepackage{microtype}      % microtypography
\usepackage{amsmath}
\usepackage[usestackEOL]{stackengine}
\usepackage{subcaption}
\usepackage{adjustbox}
\usepackage{graphicx}
\usepackage{amssymb}
\usepackage{xcolor}
\usepackage{amsmath,amsfonts,bm}

\def\eqref#1{equation~\ref{#1}}
\def\1{\bm{1}}

\def\vx{{\mathbf{x}}}

\def\vz{{\mathbf{z}}}

\def\mD{{\bm{D}}}
\def\mE{{\bm{E}}}

\DeclareMathAlphabet{\mathsfit}{\encodingdefault}{\sfdefault}{m}{sl}
\SetMathAlphabet{\mathsfit}{bold}{\encodingdefault}{\sfdefault}{bx}{n}

\usepackage{colortbl}
\usepackage{rotating}
\usepackage{mathtools}
\usepackage{tikz}
\usepackage{multirow}
\usepackage{amsmath}

\title{Nested Diffusion Models Using Hierarchical Latent Priors}

\author{\textbf{Xiao Zhang$^{*1}$ \enspace\enspace Ruoxi Jiang$^{*1}$\enspace\enspace Rebecca Willett$^{1,2}$\enspace\enspace Michael Maire$^{1}$}}

\begin{document}
\twocolumn[{%
   \renewcommand\twocolumn[1][]{#1}%
   \vspace{-28pt}
   \maketitle%
   \vspace{-10pt}
   \begin{minipage}[t]{1.0\linewidth}%
      \captionsetup{type=figure}%
      \vspace{-10pt}
      \begin{subfigure}{0.24\linewidth}
        \centering
        \includegraphics[width=\linewidth]{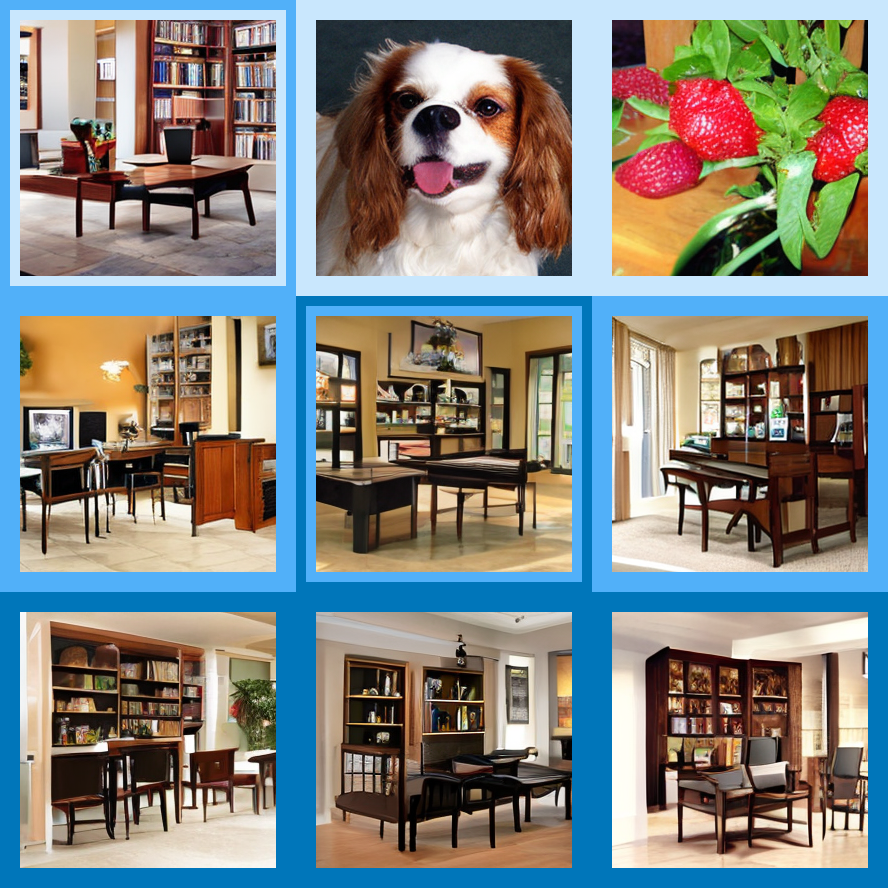}
      \end{subfigure}\hfill
      \begin{subfigure}{0.24\linewidth}
        \centering
        \includegraphics[width=\linewidth]{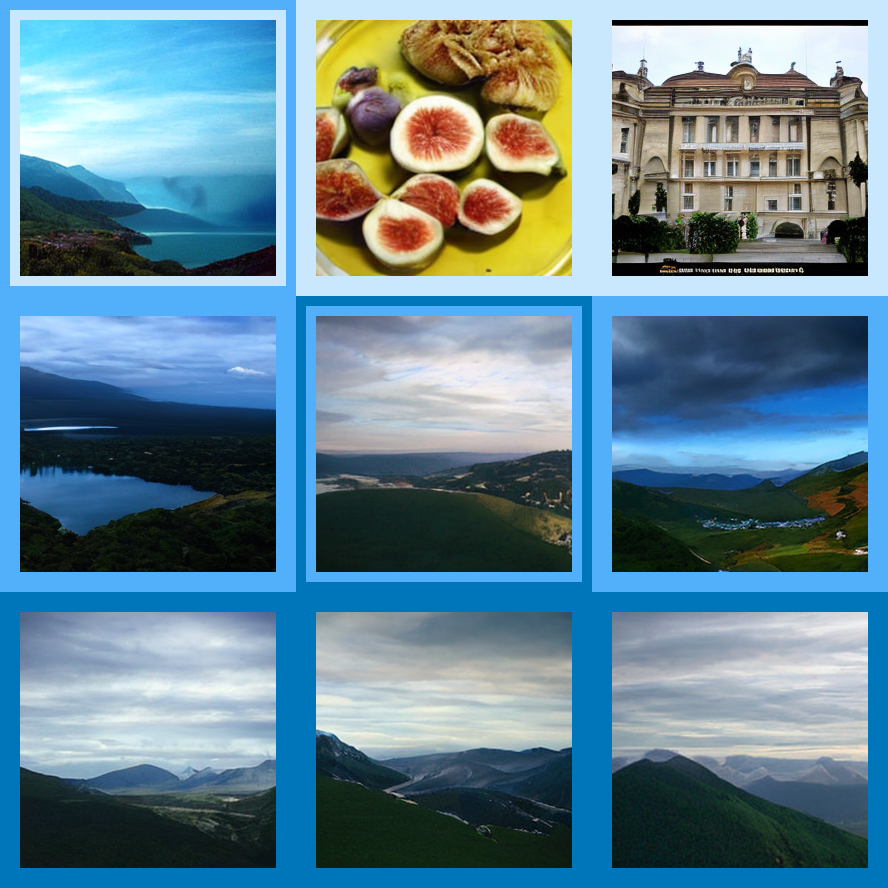}
      \end{subfigure}\hfill
      \begin{subfigure}{0.24\linewidth}
        \centering
        \includegraphics[width=\linewidth]{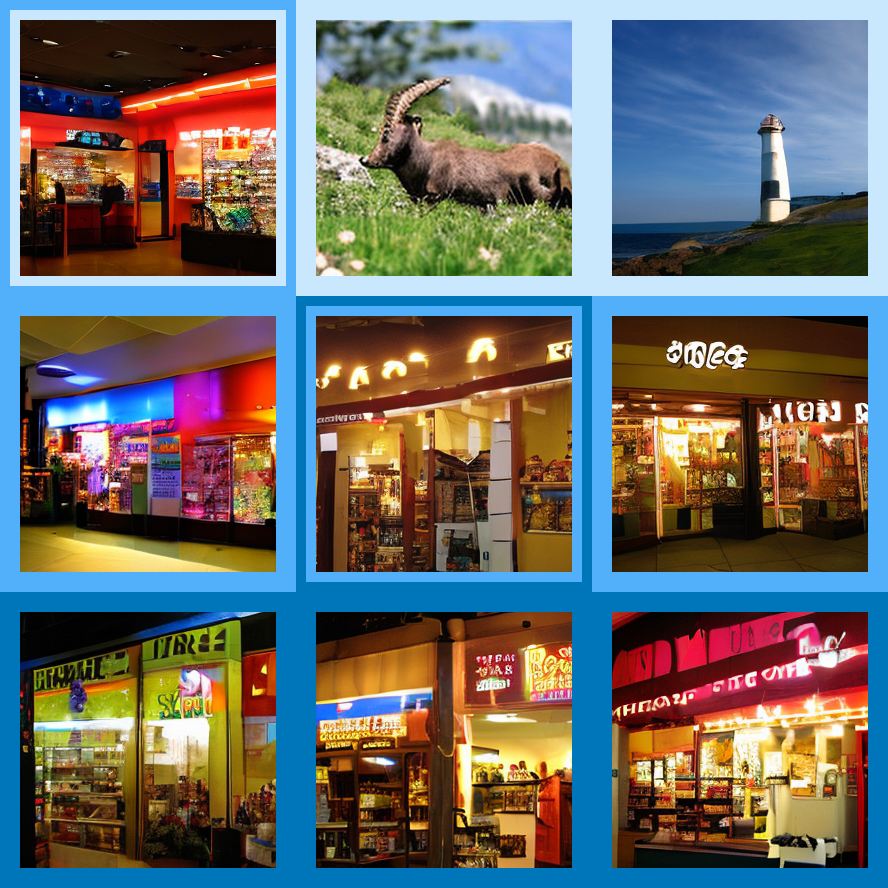}
      \end{subfigure}\hfill
      \begin{subfigure}{0.24\linewidth}
        \centering
        \includegraphics[width=\linewidth]{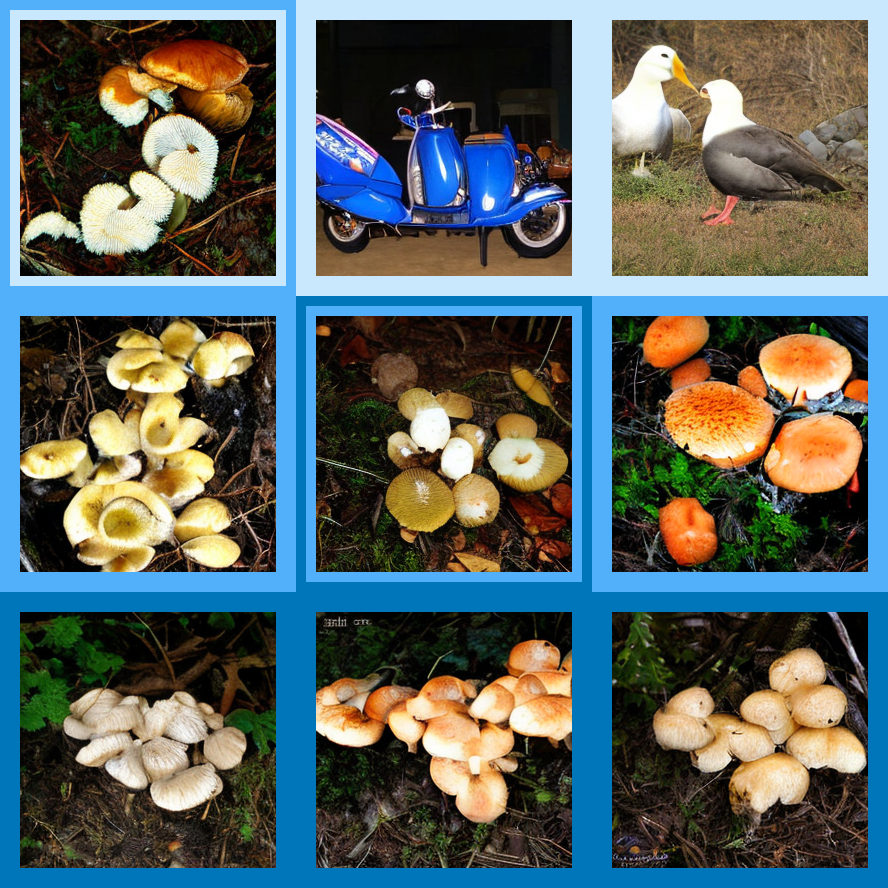}
      \end{subfigure}
    \caption{%
    \textbf{Image generation via diffusion models nested along a hierarchical semantic chain.}
    We synthesize images using a sequence of diffusion models to generate a hierarchy of latent representations, starting from a low-dimensional semantic feature embedding and refining to a detailed image.  At each hierarchical level, synthesis of a higher-dimensional latent from noise is conditioned on the more abstract latents generated at levels above.  Here, each successive row visualizes resulting images when fixing latents up to some level and resampling those at subsequent levels; images (darker background) are produced by resampling only the more detailed levels of the hierarchical representation of a specific image in the preceding row (lighter background).  Trained on ImageNet-1K, our multi-level generation system, \emph{free of any external conditioning} (\ie, no class labels), learns a hierarchy that transitions from reflecting abstract semantic similarities to fine visual details.}
      \label{fig:teaser}
   \end{minipage}%
   \vspace{15pt}
}]%

\def\thefootnote{*}\footnotetext{Equal contribution. $^1$Department of Computer Science, The University of Chicago $^2$Department of Statistics, The University of Chicago. Correspondence to: Xiao Zhang <zhang7@uchicago.edu>, Ruoxi Jiang <roxie62@uchicago.edu>.} \def\thefootnote{\arabic{footnote}}
\begin{abstract}
We introduce nested diffusion models, an efficient and powerful hierarchical generative framework that substantially enhances the generation quality of diffusion models, particularly for images of complex scenes. Our approach employs a series of diffusion models to progressively generate latent variables at different semantic levels. Each model in this series is conditioned on the output of the preceding higher-level models, culminating in image generation. Hierarchical latent variables guide the generation process along predefined semantic pathways, allowing our approach to capture intricate structural details while significantly improving image quality. To construct these latent variables, we leverage a pre-trained visual encoder, which learns strong semantic visual representations, and modulate its capacity via dimensionality reduction and noise injection. Across multiple datasets, our system demonstrates significant enhancements in image quality for both unconditional and class/text conditional generation. Moreover, our unconditional generation system substantially outperforms the baseline conditional system. These advancements incur minimal computational overhead as the more abstract levels of our hierarchy work with lower-dimensional representations.
\end{abstract}
\section{Introduction}
\begin{figure}[h]
    \centering
    \begin{subfigure}{0.5\linewidth}
    \includegraphics[width=\linewidth]{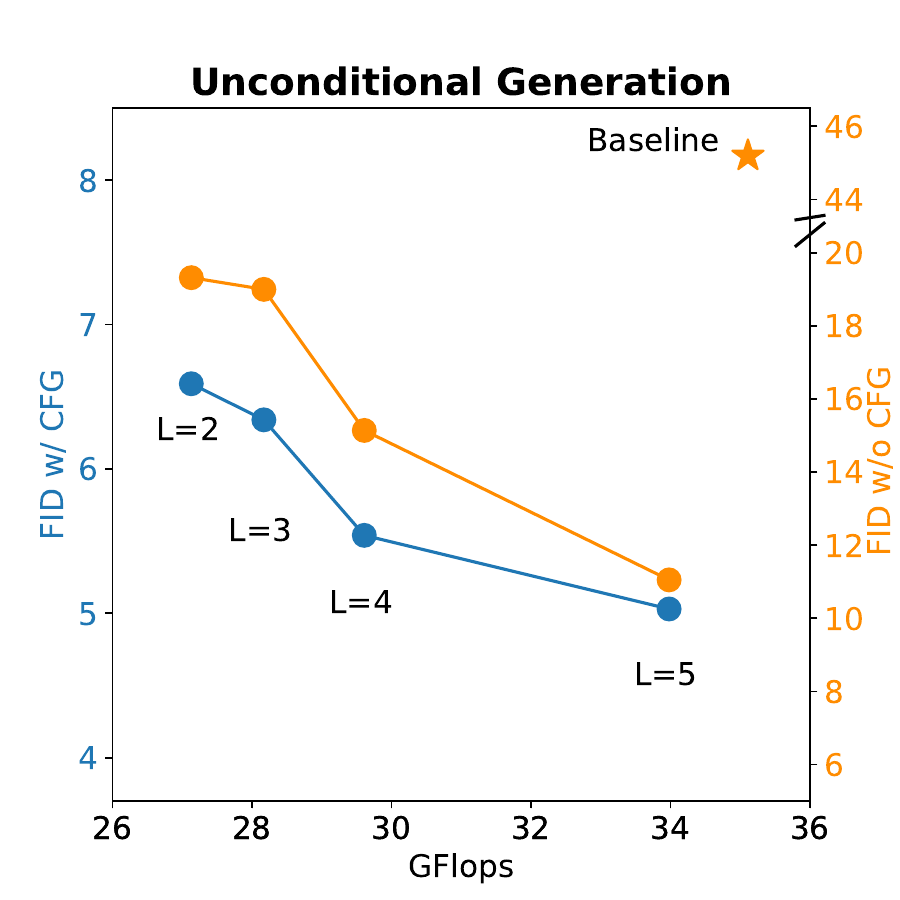}
    \end{subfigure}\hfill
    \begin{subfigure}{0.5\linewidth}
    \includegraphics[width=\linewidth]{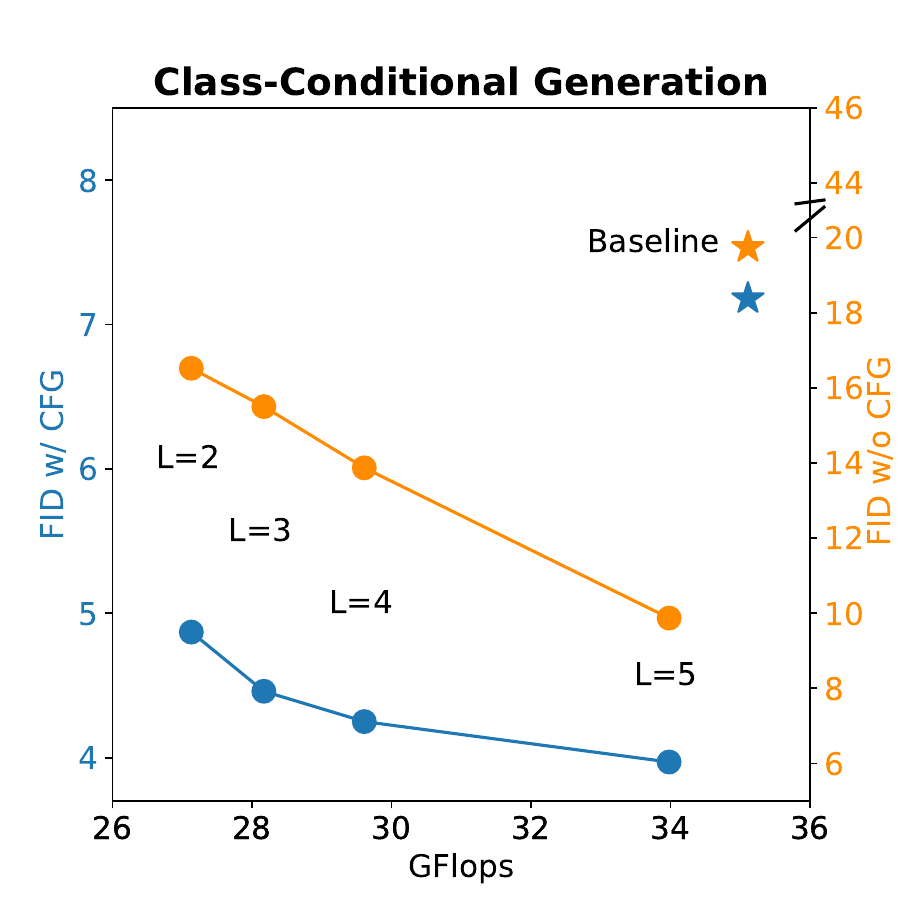}
    \end{subfigure}
    \caption{\textbf{Image generation quality when scaling our nested diffusion models on ImageNet-1K dataset.} 
    The deeper hierarchies we build lead to a slight increase in computational overhead (particularly when $L \leq 4$), as measured by GFlops, while significantly improving the generation quality.
    Compared to the single-level baseline model using comparable GFlops, our 5-level unconditional system significantly improves the performance w/o classifier-free guidance (CFG) by reducing FID from 45.19 to 11.05, bypassing the class-conditional baseline of 19.74.}
    \label{fig:scale-gflops}
\end{figure}
\begin{figure*}[t!]
    \centering
\includegraphics[width=1.0\textwidth, trim={2cm, 2cm, 6.5cm, 2cm}, clip]{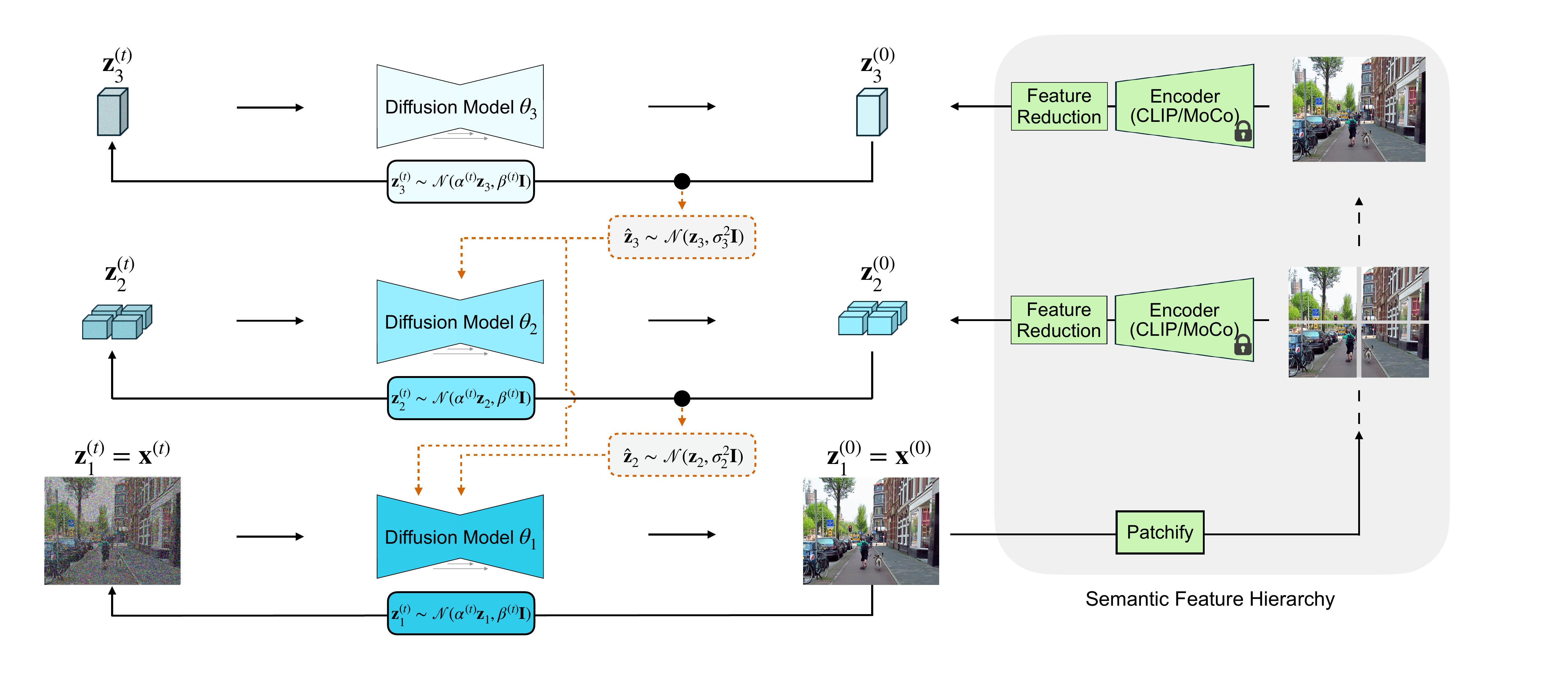}
    \caption{\textbf{Nested diffusion architecture.}
    \textit{\textbf{Left:}}
    We train a sequence of diffusion models to generate a hierarchical collection of latent representations $\{\vz_3, \vz_2, \vz_1 = \vx\}$ of increasing dimensionality up to an image $\vz_1 = \vx$.  Generated latents serve as conditional inputs (dotted lines) to diffusion models at subsequent levels, with separately parameterized noising processes, $\hat{\vz}_{l}\sim \mathcal{N}(\vz_l, \sigma_{l}^2\mathbf{I})$, controlling the information capacity of these  signals.
    %We employ a sequence of diffusion models to generate target latent representations, ultimately producing final images. The arrows with solid lines indicate the generative (backward) process for each diffusion model, aimed at generating latent features at the corresponding hierarchy level. Dotted lines indicate how latent features from higher hierarchies are injected into diffusion models at lower levels.
    \textit{\textbf{Right:}}
    A pre-trained, frozen visual encoder provides target latent representations for each level of the hierarchy. To construct these latent features, we run the encoder on patchified images, reducing patch size and applying dimensionality reduction across feature channels in order to shift focus from local details to global semantics. Upper level targets encode more abstract semantics and, being lower-dimensional vectors, are less computationally expensive to synthesize, making hierarchical generation fast.}%
    \label{fig:diagram}
\end{figure*}

The modern era of computer vision opened with deep networks driving advances in representation learning: mapping images, patches, or pixels to feature vectors that encode semantic information and support a range of downstream tasks such as classification~\cite{DBLP:conf/nips/KrizhevskySH12, DBLP:journals/corr/SimonyanZ14a, DBLP:conf/cvpr/HeZRS16}, segmentation~\cite{DBLP:conf/cvpr/LongSD15, DBLP:conf/cvpr/HariharanAGM15}, and object detection~\cite{DBLP:journals/pami/GirshickDDM16, DBLP:conf/iccv/HeGDG17, DBLP:conf/eccv/CarionMSUKZ20}. A variety of deep generative methods have since emerged to enable the reverse mapping, from a given prior or a learned embedding, back to the space of images. GANs~\citep{goodfellow2014generative}, VAEs~\citep{kingma2013auto, sonderby2016ladder, vahdat2020nvae, pervez2020variance, luhman2022optimizing}, normalizing flows~\citep{papamakarios2021normalizing, abdal2021styleflow, wang2022low}, and diffusion models~\citep{gu2022vector, gu2023matryoshka, zhang2023adding, song2020denoising} have demonstrated capacity to synthesize complex real-world image, video, and language data~\citep{bao2023all, nichol2021glide, liu2024alleviating}. In parallel, representation learning has advanced through development of scalable architectures and training objectives, yielding self-supervised approaches, including contrastive learning~\cite{chen2021empirical, chen2020simple, he2020momentum, caron2021emerging}, masked autoencoders (MAEs)~\cite{he2022masked}, and hybrids~\cite{zhou2021ibot}, that rival supervised feature learning.

Although generation and representation learning may have different immediate applications, they are inherently linked. A process that synthesizes realistic images must internally capture some notion of semantics in order to produce globally coherent structure. Indeed, recent work investigating generative models reveals that they capture rich visual representations useful for downstream tasks: segmentation~\cite{baranchuk2021label,xu2023open}, image intrinsics~\cite{du2023generative} and image recognition~\cite{yang2023diffusion,li2023your}. Conversely, another branch of research demonstrates that strong visual representations can further enhance generation quality via: conditioning on clustered features~\cite{hu2023self}, learning to generate visual features that serve as a conditional signal~\cite{li2023self}, or adding a self-supervised representation learning loss to a generative model~\cite{li2023mage,yu2024representation}. However, these uses of pre-trained visual encoders or feature learning objectives focus only on abstract, high-level features, and in essence may function as unsupervised substitutes for image category labels.

Images contain diverse, multi-scale structures, from local textures and edges to parts, objects, and coherent scenes. For a generative system to produce realistic images, it must model all of these aspects. Current generative models frequently struggle to accurately represent attributes such as physical properties~\cite{kang2024far} and geometric layout~\cite{sarkar2024shadows}, suggesting that conventional generative training objectives are insufficient for capturing these complex visual relationships. 

To address this, we anchor a generation process to a visual feature hierarchy, which provides intermediate targets to guide progressive image synthesis. Our system employs a series of diffusion models, each operating at a different level of semantic abstraction and conditioned on outputs from higher levels. We build training targets for this hierarchical generator using a pre-trained visual encoder, applied to image patches of varying scale, in order to represent visual structures ranging from local texture to global shape. As additional controls on our target feature hierarchy, we compress feature representations through dimensionality reduction and noise-based perturbation. These capacity controls are essential to prevent memorization of image details at intermediate levels, allowing us to scale our model to deep hierarchies.

Unlike traditional methods using VAE features~\cite{rombach2022high} or image pyramids~\cite{gu2023matryoshka} that primarily focuses on local textures, our approach emphasizes structured, multi-scale semantic representations. Compared to the hierarchical VAE~\cite{vahdat2020nvae, zhao2017learning, child2020very, takida2023hq} models that run generation in a hierarchical latent space but suffer training instability, we use frozen latent representations, which significantly enhances training stability and yields much better generation quality.

Figure~\ref{fig:teaser} shows example output using our method for unconditional synthesis on ImageNet-1k. Our unconditional system even outperforms the conditional generation baseline, as benchmarked in Figure~\ref{fig:scale-gflops}, and also achieves consistent quality improvement as the number of levels $L$ in the hierarchy increases from $2$ to $5$. Figure~\ref{fig:diagram} sketches the key components of our model architecture. Section~\ref{sec:experiments} extends experiments to the challenging setting of text-conditioned image generation trained on COCO scenes, where our hierarchical model outperforms baseline models containing substantially more parameters and consuming significantly larger training datasets. Our contributions as follows:
\begin{itemize}
    \vspace{2pt}%
    \item We introduce nested diffusion models, anchoring image generation to a hierarchical feature representation. Top hierarchy levels promote consistency in global image structure, while subsequent levels refine visual details. Resampling specific levels gives tunable control over synthesis.%
    \vspace{2pt}%
    \item Our design greatly enhances generation quality while maintaining efficiency. Our five-level hierarchical model increases computational cost, measured in GFlops, by only $25\%$ relative to single-level diffusion, yet significantly improves quality. Relative to a baseline model requiring comparable GFlops, we decrease FID from 45.19 to 11.05 for unconditional generation and from 31.13 to 9.87 for conditional generation on ImageNet-1k.
    \vspace{2pt}%
    \item Our system consistently improves performance in both conditional and unconditional generation tasks as more hierarchical levels are added. Notably, on ImageNet-1k, our unconditional generation quality surpasses that of the baseline class-conditional diffusion model.
\end{itemize}

\section{Related Work}

\textbf{Hierarchical models.}
Hierarchical variational autoencoders (HVAEs) \citep{vahdat2020nvae, zhao2017learning, child2020very, takida2023hq} extend the latent space of VAEs~\citep{kingma2013auto} to include multiple variables, and demonstrate improved generation quality. However, HVAEs are known to suffer from high variance and collapsed representations, where the top-level variables may be ignored \citep{vahdat2020nvae, child2020very}. To address this issue, \citet{luhman2022optimizing} introduce a layer-wise scheduler and regularization to enhance stability, while \citet{hazami2022efficientvdvae} propose a simplified architecture.

Recent work has sought to build hierarchical generative systems by freezing the latent variables and leveraging powerful generative methods such as diffusion models and autoregressive models. For example, \citet{ho2022cascaded, gu2023matryoshka, liu2024alleviating} train a set of diffusion models to handle images at different resolutions, and \citet{tian2024visual} train a hierarchical autoregressive model to predict the residuals between tokenized representations at adjacent resolutions. However, none of these approaches involve training with hierarchical semantic latent representations.

\textbf{Conditional generation.}
A conditional diffusion model aims to parameterize the prior as a complex joint distribution conditioned on an input, rather than using a simple Gaussian prior, significantly enhancing the model’s capacity to capture intricate data patterns. For images of complex scenes, generation conditioned on image captions \cite{gu2022vector, kang2023scaling, reed2016generative} has shown notable improvements in both quality and controllability. \cite{zhang2023adding, rombach2022high} extend this conditioning approach to multiple modalities, incorporating input such as segmentation, depth maps, and human joint positions. Another direction in this field is learning the conditional variable itself. Models like DiffAE~\citep{preechakul2022diffusion}, SODA~\citep{hudson2024soda}, and \citet{abstreiter2021diffusion} train an encoder to produce a low-dimensional latent variable to assist the generation process; these works also demonstrate that such an encoder can learn meaningful image representations.

\textbf{Generation with semantic visual representations.}
State-of-the-art generative models, such as diffusion and autoregressive models, can be viewed as denoising autoencoders that inherently learn meaningful data representations. \citet{yang2023diffusion, tang2023emergent, zhang2024deciphering} demonstrate that diffusion models capture semantic visual representations, which are directly applicable to various downstream tasks~\citep{baranchuk2021label,karazija2023diffusion}. \citet{zhang2023structural} highlight that a discriminator in a GAN can learn useful representations. \cite{li2023mage, jiang2024training} show that incorporating representation learning objectives into the generative framework can further enhance generation quality. \citet{li2023self, hu2023self, wang2024instancediffusion} leverage semantic representations learned by the encoder to further improve generation quality.
\section{Method}
We employ a structured approach to capture hierarchical semantic features for image generation.
%Here, we review diffusion models, one essential component of our system.

\subsection{Preliminary: Diffusion models}
As a generative framework, diffusion models \citep{ho2020denoising, song2020score, song2020denoising} consist of both a forward (diffusion) process and a backward process, each spanning over $T$ steps.
Let $\mathbf{x} \in \mathbb{R}^{d}$ denote the original data sample.
The forward process defines a sequence of latent variables $\{\vx^{(t)}\}_{t=1}^T$ obtained by sampling from a Markov process parameterized as $q\left(\vx^{(t)} \mid \vx^{(t-1)} \right):= \mathcal{N}(\vx^{(t)}; \alpha^{(t)}\vx, {\beta^{(t)}}\mathbf{I})$, where $\alpha^{(t)}$ and $\beta^{(t)}$ are hyperparameters of the noise scheduler, ensuring that the signal-to-noise ratio (SNR) decreases as $t$ increases.

In the backward process, the model $\mD_{\theta}$ is tasked with estimating the transition probability $p(\vx^{(t-1)}|\vx^{(t)})$ and generating data through the process $\prod_{t=1}^T p_{\theta}(\vx^{(t-1)}|\vx^{(t)})p(\vx^{(T)})$, where $p_{\theta}(\vx^{(t-1)}|\vx^{(t)})$ represents the transition probability estimated by $\mD_{\theta}$. It is trained by maximizing the Variational Lower Bound (VLB) \citep{kingma2021variational}:
\begin{equation}
  \mathcal{L}_{\rm VLB} =  
  - \sum_{t=1}^T D_{\rm KL}\left(q \left(\vx^{(t-1)}|\vx^{(t)}, \vx \right) \middle\| p_{\theta}\left(\vx^{(t-1)}|\vx^{(t)}\right)\right).
  \label{eqn:vlb}
\end{equation}
Here $q\left(\vx^{(t-1)}|\vx^{(t)}, \vx \right)$ can be derived using Bayes’ rule as $q\left(\vx^{(t)}|\vx^{(t-1)}, \vx \right)q\left(\vx^{(t-1)}|\vx \right)/q\left(\vx^{(t)}|\vx \right)$. 
Maximizing RHS of Eqn.\ref{eqn:vlb} can be simplified as training $\mD_{\theta}$ to estimate the noise $\boldsymbol{\epsilon}^{(t)} \in \mathbb{R}^d$ sampled from $\mathcal{N}(0,\mathbf{I})$ \citep{ho2020denoisingdiffusionprobabilisticmodels}:
\begin{equation*}
  \mathcal{L}_{\rm{diffusion}} = \mathbb{E}_{\boldsymbol{\epsilon}^{(t)}, t} \|\mD_{\theta}(\alpha^{(t)}\vx  + \beta^{(t)}\boldsymbol{\epsilon}^{(t)}, t) - \boldsymbol{\epsilon}^{(t)}\|_2.
\end{equation*}

\subsection{Nested diffusion models}
We propose a hierarchical generative framework with $L$ levels, where each level is instantiated as a diffusion model $\mD_{\theta_l}$.
As illustrated in Figure~\ref{fig:diagram}, the generative model at level $l$ produces latent variable $\vz_l \in \mathbb{R}$ with $p_{\theta_l}(\vz_l|\vz_{>l})$, with $\vz_{>l} := \{\vz_m\}_{m>l}$ representing latent variables from higher levels. 
The feature dimension of $\vz_l \in \mathbb{R}^{d_l}$ decreases as $l$ increases, such that $d_l > d_{l+1}$.
At the shallowest level when $l=1$, the latent variables correspond directly to the data samples, that is, $\mathbf{z}_1 = \mathbf{x}$. 

\textbf{Diffusion with semantic hierarchy.} Our approach explicitly guides the generative process to align with a semantic hierarchy. Here, the top tier (larger $l$) denotes higher levels of semantic abstraction, whereas the lower tier (smaller $l$) represents detailed, fine-grained information.  
This is essential for preserving image semantic structures and producing realistic samples in generative models. 

\textbf{Non-Markovian generation.}
At each hierarchical level $l$, we follow the diffusion model framework and task $\mD_{\theta_l}$ to estimate the transition probability $p_{\theta_l}(\vz^{(t-1)}_{l}|\vz^{(t)}_{l}, \vz_{>l})$. 
At layer $l$, we assume a non-Markovian generation process where $\mD_{\theta_l}$ depends on the entire set of latent variables $\vz_{>l}$ estimated from the preceding hierarchies.

We optimize the hierarchal diffusion models using the following objectives:
\begin{align}
  \label{eqn:nested_diffusion_elbo}
  &\mathcal{L}_{\rm hierarchical\_ELBO} = \\ &\sum_{l=1}^{L-1}  
  \sum\limits_{t=1}^{T}  D_{\rm KL}\left(q \left(\vz_l^{(t-1)}|\vz_l^{(t)}, \vz_{l}, \vx \right) \middle\| p_{\theta_l}\left(\vz_l^{(t-1)}|\vz_l^{(t)}, \vz_{>l}\right)\right) \nonumber \\
   &+  \sum\limits_{t=1}^T D_{\rm KL}\left(q \left(\vz_{L}^{(t-1)}|\vz_{L}^{(t)}, \vz_{L}, \vx \right) \middle\| p_{\theta_l}\left(\vz_{L}^{(t-1)}|\vz_{L}^{(t)}\right)\right). \nonumber
\end{align} 

Comparing to hierarchical VAEs which also includes hierarchical latent variables $\{\vz_l\}_{l=1}^L$, we enhance sampling capability by integrating the diffusion model and introducing an additional set of latent variables $\{\vz_l^{(t)}\}_{t=0}^T$ for each level $l$. 
This modification allows for multiple sampling steps, as opposed to the single forward pass used in hierarchical VAEs, leading to a more accurate prior estimation. 
This improvement is vital in hierarchical generative systems, where mismatches between the posterior and prior distributions can compound across levels, potentially degrading the quality of the generated output.

\begin{figure}[t]
    \centering
    \begin{minipage}{0.49\linewidth}
    \begin{minipage}[t]{0.33\linewidth}
    \centering
    \includegraphics[width=\linewidth]{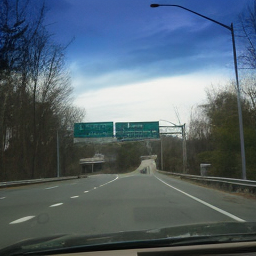}
    {\small Image}
    \end{minipage}\hfill
    \begin{minipage}[t]{0.33\linewidth}
    \centering
    \includegraphics[width=\linewidth]{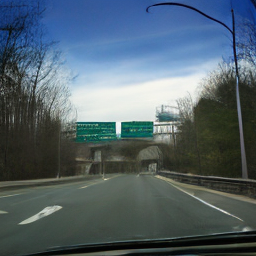}
    {\small $\sigma_2 = 0$ }
    \end{minipage}\hfill
    \begin{minipage}[t]{0.33\linewidth}
    \centering
    \includegraphics[width=\linewidth]{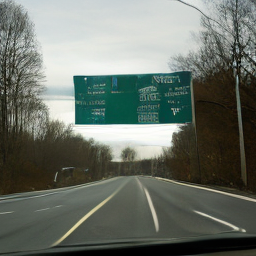}
    {\small $\sigma_2 = 0.5$ }
    \end{minipage}
    \end{minipage}\hfill
    \begin{minipage}{0.49\linewidth}
    \begin{minipage}[t]{0.33\linewidth}
    \centering
    \includegraphics[width=\linewidth]{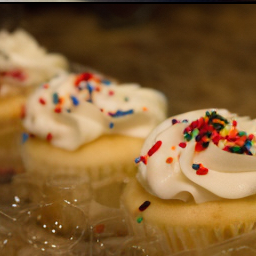}
    {\small Image}
    \end{minipage}\hfill
    \begin{minipage}[t]{0.33\linewidth}
    \centering
    \includegraphics[width=\linewidth]{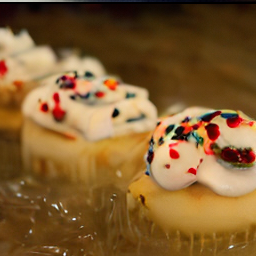}
    {\small $\sigma_2 = 0$ }
    \end{minipage}\hfill
    \begin{minipage}[t]{0.33\linewidth}
    \centering
    \includegraphics[width=\linewidth]{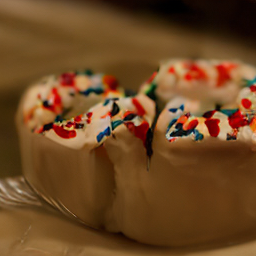}
    {\small $\sigma_2 = 0.5$ }
    \end{minipage}
    \end{minipage}
    \caption{\textbf{Feature compression via Gaussian noise.}
    For a two-level hierarchical generator ($L=2$), we generate images conditioned on an oracle CLIP feature $\vz_2$, inferred from input images, with feature channels reduced from 512 to 256 dimensions via SVD. Without noise ($\sigma_2 = 0$) added to $\vz_2$, the generator $\mD_{\theta_1}$ degenerates to an autoencoder that nearly reconstructs the input; adding Gaussian noise ($\sigma_2 = 0.5$) to $\vz_2$ limits feature information, allowing for generation of new content.}
    \label{fig:recon_img}
\end{figure}
\subsection{Hierarchical latents \& progressive compression}
\label{sec:latent_variable}

\textbf{Extraction of hierarchical features.}  
We construct $\{\vz_l\}_{l=1}^{L}$ from pre-trained visual encoder and keep those frozen during training. 
Most visual encoders, denoted as $\mE(\vx) = \vz$, map an input image $\vx$ to a vector $\vz$. 
To build a hierarchical representation, we propose running $\mE$ on image patches. Let $C(\vx, M) = \{\vx_m\}_{m=1}^{M^2}$ represent the cropping operation that splits the image $\vx \in \mathbb{R}^{H \times W \times 3}$ into $M^2$ non-overlapping square patches $\vx_m \in \mathbb{R}^{\frac{H}{M} \times \frac{W}{M} \times 3}$. We build a latent variable $\vz_l = \mE\left(C\left(\vx, L - l + 1\right)\right)$ over image patches. As patch size decreases, the visual representation transitions from capturing global structures to more localized features.

\textbf{Feature compression via Gaussian noise.} 
 Our construction of $\{\vz_l\}_{l =1}^{L}$ contrasts with hierarchical VAEs, which use compression objectives to learn hierarchical latent variables but often face high variance issues, as noted in prior work~\citep{pervez2020variance, vahdat2020nvae, child2020very}. While we address instability, our design presents a new challenge: representations from the visual encoder tend to be highly informative, allowing the generative model to reconstruct the input image accurately, which can cause the generator to behave like an autoencoder.

We show this effect in Figure~\ref{fig:recon_img}: when conditioning on the oracle CLIP visual features, diffusion model could nearly reconstruct the input: $p_{\theta_{L}}(\vz_0|\vz_{L}) \approx 1$, essentially “bypassing” the middle levels $p_{\theta_{l}}(\vz_{l}|\vz_{l+1})$ in a hierarchical system.
Consequently, we must reduce the information contained in $\vz_l$
to ensure each hierarchical level contributes meaningfully to the generation process and our procedures are as follows:

\textbf{Channel reduction via singular value decomposition.} In our patch-based approach, the channel number of the latent variable quadratically increases as we move down the hierarchy, with $\vz_l \in \mathbb{R}^{(L - l + 1)^2 \times d}$, quickly making the information overcomplete for generation. To address this, we propose trimming the feature channels. Specifically, we apply singular value decomposition (SVD) to the encoder’s feature vector, preserving only the leading $d / (L - l + 1)$ channels, resulting in 
$\vz_l \in \mathbb{R}^{(L - l + 1) \times d}$, with channel linearly increased over level of hierarchy. 

\textbf{Information reduction through Gaussian noise.} 
As shown in Figure~\ref{fig:recon_img}, channel reduction alone is insufficient to prevent the diffusion model from degrading into an autoencoder. To further increase the abstraction level, we introduce Gaussian noise to $\vz_l$, represented as $\hat{\vz}_{l}\sim \mathcal{N}(\vz_l, \sigma_{l}^2\mathbf{I})$, where $\sigma_l$ is a fixed constant based on the hierarchical level.
This process limits the amount of information that can be transmitted, measured by the KL divergence  $D_{KL}\left(\mathcal{N}\left(\vz_l, \sigma_l^2\right), \mathcal{N}\left(\mathbf{0}, \mathbf{I}\right)\right)$. 
A large variance $\sigma_l^2$ substantially limits the information capacity. In our experiments, adding Gaussian noise proved essential in preserving and enhancing generation quality as the number of hierarchical levels increased. We only add Gaussian noise during training and remove it during generation.
  
With these approaches, our loss functions are as follows:
\begin{align}\label{eqn:nested_diffusion}
&\mathcal{L}_{\rm{nested\_diffusion}} =  
\\ 
&\sum_{l=1}^{L-1}\mathbb{E}_{\hat{\vz}_{>l},  \boldsymbol{\epsilon}^{(t)}_l, t} \|\mD_{\theta_l}(\alpha^{(t)}\vz_l  + \beta^{(t)}\boldsymbol{\epsilon}^{(t)}_l, \hat{\vz}_{>{l}}, t) - \boldsymbol{\epsilon}^{(t)}_l\|_2  \nonumber 
\\
& + \mathbb{E}_{\boldsymbol{\epsilon}^{(t)}_l, t} \|\mD_{\theta_{L}}(\alpha^{(t)}\vz_{L}  + \beta^{(t)}\boldsymbol{\epsilon}^{(t)}_l , t) - \boldsymbol{\epsilon}^{(t)}_l\|_2, \nonumber
\end{align}

where $\boldsymbol{\epsilon}^{(t)}_l \in \mathbb{R}^{d_l}$ denotes noise sampled at each level.

\subsection{Diffusion with semantic consistent neighbors}
\label{sec:semantic_neighbours}
\begin{figure}[t]
\small
    \centering
\begin{minipage}{1.0\linewidth}
\begin{minipage}{0.19\linewidth}
    \centering
    {\footnotesize Input Image}
    \textcolor{black}{\rule[1ex]{\linewidth}{1pt}}
    \vspace{-20pt}
\end{minipage}\hfill
\begin{minipage}{0.39\linewidth}
    \centering
    {\footnotesize Neighbored Images with CLIP Features}
    \textcolor{black}{\rule[1ex]{\linewidth}{1pt}}
\end{minipage}\hfill
\begin{minipage}{0.39\linewidth}
    \centering
    {\footnotesize Neighbored Images with VAE Features}
    \textcolor{black}{\rule[1ex]{\linewidth}{1pt}}
\end{minipage}

\begin{minipage}{0.19\linewidth}
\includegraphics[width=1.0\textwidth]{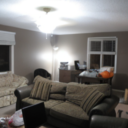}
\end{minipage}\hfill
\begin{minipage}{0.39\linewidth}
\centering
\includegraphics[width=1.0\textwidth]{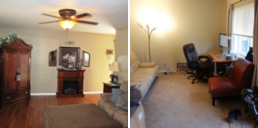}
\end{minipage}\hfill
\begin{minipage}{0.39\linewidth}
\includegraphics[width=1.0\textwidth]{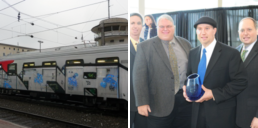}
\end{minipage}\hfill

\begin{minipage}{0.19\linewidth}
\includegraphics[width=1.0\textwidth]{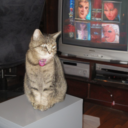}
\end{minipage}\hfill
\begin{minipage}{0.39\linewidth}
\centering
\includegraphics[width=1.0\textwidth]{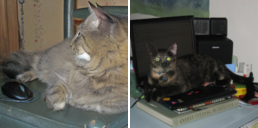}
\centering
\end{minipage}\hfill
\begin{minipage}{0.39\linewidth}
\includegraphics[width=1.0\textwidth]{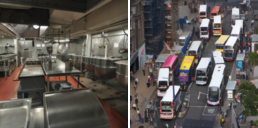}
\centering
\end{minipage}\hfill
%\vspace{4pt} 
%\begin{tikzpicture}
%\hspace{10pt}
%$    \begin{scope}
%$      \draw[->, thick](2,5) -- (12,5) node[pos=0.5,below]{\scriptsize Decreasing feature dimensions and increasing semantic abstraction levels};
%$    \end{scope}
%\end{tikzpicture}
\end{minipage}
\caption{\textbf{Visualization of K-Nearest Neighbors (KNN) with different sources of latent features.} For each input image, we display neighboring images,  based on features extracted from two types of visual representations: CLIP representations, and VAE bottlenecks. Unlike the VAE, which focuses on low-level visual structures, CLIP emphasizes semantic representations, yielding more meaningful nearest neighbors. Our experiments demonstrate that running a diffusion model on a latent space with well-structured neighbors is essential for enhancing generation quality.}
\label{fig:knn}
\end{figure}

Diffusion model is highly related to the \textit{mean-shift} iterations ~\cite{comaniciu1999mean, cheng1995mean}, as highlighted by the analysis~\cite{wang2023score, song2020improved, karras2022elucidating}. Specifically, assuming hierarchical  dependency $p(\vz_l^{(t)}|\vz_{>l}) = \mathbb{E}_{\vz_l}p(\vz_l^{(t)}|\vz_l, \vz_{>l}) = \mathbb{E}_{p(\vz_l|\vz_{>l})}p(\vz_l^{(t)}|\vz_l)$, an optimal denoiser $\mD^{*}_{l}$ for Eqn. \ref{eqn:nested_diffusion} can be expressed in closed-form as follows~\cite{song2020improved,karras2022elucidating}:
\begin{equation}
    \mD^{*}_{l}(\vz_l^{(t)}|\vz_{>l}) = \frac{\sum_{\vz_l}p(\vz_l^{(t)}|\vz_l)p(\vz_l|\vz_{>l})\vz_l}{\sum_{\vz_l}p(\vz_l^{(t)}|\vz_l)p(\vz_l|\vz_{>l})}.
    \label{eqn:optimal_denoiser}
\end{equation}
This suggests that the optimal solution is the weighted data point $\vz_l$, based on the similarity between $\vz_l$ and $\vz_l^{(t)}$. Therefore, the structures of $\vz_l$ significantly impact the quality of optimal denoiser. Ideally, the neighbor of both $\vz_l^{(t)}$ and $\vz_l$ should have similar semantic structures to ensure robust mean-shift iterations. In Figure~\ref{fig:knn}, we attempt to visualize the structure of $\vz_l$ via nearest neighbor images with CLIP features or VAE features. Unlike VAE, which focuses on low-level textures and results in unrelated neighbored images, CLIP yields semantically similar images and better generation quality in our experiments.
    
\section{Experiments}
\label{sec:experiments}

We present the setup and results of our experiments, where we evaluate the performance of our nested diffusion model across various tasks. Our primary focus is to explore the model’s effectiveness in both conditional and unconditional image generation scenarios using the COCO-2014\citep{lin2014microsoft} and ImageNet-1K datasets~\citep{russakovsky2015imagenet}.

\subsection{Experimental Setup}

\textbf{Nested diffusion models.} We utilize U-ViT~\citep{bao2023all}, a ViT-based UNet model with an encoder-decoder architecture, for $\mD_{\theta_l}$. 
This model employs skip connections and performs diffusion in the latent space of a pre-trained VAE, reducing the input size from $256\times256\times3$ to $32\times32\times4$, which enables efficient handling of high-resolution images. 
We customize the network configurations to make it aligned with the standard ViT-Base model: The transformer model we use consists of 12 blocks, with the base channel dimension set to 768, and each attention block comprises 12 attention heads. 
We use the default diffusion scheduler, sampler, and training hyperparameters from U-ViT~\citep{bao2023all} for ImageNet-1k and COCO respectively.

To construct our nested diffusion model, we use the same network configuration for each hierarchical level. At higher hierarchical levels, there is a progressive reduction in the dimensionality of $\vz_l$, which leads to minimal extra computational cost even though the number of parameters increases. 

\begin{table}[t]
\small
\centering
\strutlongstacks{T}
\centering
\begin{minipage}{\linewidth}
\begin{subtable}{\linewidth}
\centering\begin{tabular}{l|ccc|ccc}
& \multicolumn{3}{c}{w/o CFG} & \multicolumn{3}{c}{w/ CFG}\\
\hline
Model & $\sigma_{2}$ = 0.0 & 0.5 & 1.0 & $\sigma_{2}$ = 0.0 & 0.5 & 1.0\\
\hline
$L = 1$ & 55.41 & - & - & - & - & - \\
$L = 1^*$ & 45.19 & - & - & - & - & - \\
\hline
$L = 2$ & \textbf{19.32} & 20.66  &  27.40 & \textbf{6.59} & 7.19 & 8.69\\
$L = 3$ &  20.34 & \textbf{19.00} & 23.37  & 6.77 & \textbf{6.34} & 6.98\\ 
$L = 4$ &  17.67 & \textbf{15.14} & 16.27 & 5.79 &\textbf{5.54}& 5.89\\
$L = 5$ &  19.04 & 11.88 & \textbf{11.05} & 7.68 & 5.36 & \textbf{5.03}\\
\hline
\end{tabular}
\caption{\textit{Unconditional} image generation for ImageNet-1k, 256$\times$256}
\label{table:uncond_in1k}
\vspace{1em}
\end{subtable}
\begin{subtable}{\linewidth}
\centering\begin{tabular}{l|ccc|ccc}
& \multicolumn{3}{c}{w/o CFG} & \multicolumn{3}{c}{w/ CFG}\\
\hline
Model & $\sigma_{2}$ = 0.0 & 0.5 & 1.0 & $\sigma_{2}$ = 0.0 & 0.5 & 1.0\\
\hline
$L = 1$ & 31.13 & - & - & 13.75 & - & - \\
$L = 1^*$ & 19.74 & - & - & 7.18 & - & - \\
\hline
$L = 2$ & \textbf{16.56} & \textbf{16.52} & 22.43 & \textbf{4.87} & 5.31 & 6.49 \\
$L = 3$ & \textbf{15.51} & \textbf{15.50} & 16.35 &  \textbf{4.46} & 4.69 & 5.15\\
$L = 4$ &17.72 & 14.38& \textbf{13.87} & 4.81 & 4.38 & \textbf{4.25} \\ 
$L = 5$ & 18.04 & 11.28 &  \textbf{9.87} & 4.26 & 4.05 & \textbf{3.97}\\
\hline
\end{tabular}
\caption{\textit{Conditional} image generation for ImageNet-1k, 256$\times$256}
\label{table:cond_in1k}
\end{subtable}
\end{minipage}
\caption{We evaluate image generation quality using the Fréchet Inception Distance (FID) on the ImageNet-1k and we report the results w/o and w/ classifier free guidnace (CFG). We benchmark our model across different noise levels and network depths $L$. To determine the noise levels $\{\sigma_l\}_{l=2}^L$, we use a top-down, greedy searching: for a model with depth $L$, we retain the optimal values $\{\sigma_l\}_{l=3}^L$ from the shallower model $<L$ and only tune the newly added level, $\sigma_2$. The generation quality improves with $L$ increases and adding Gaussian noise is crucial for better performance for deeper model. For comparison, we also provide baseline results for $L=1^*$, a single-level model with increased parameters to match the GFLOPs of $L=5$.}
\label{table:big_table}
\end{table}

To incorporate conditional features from higher hierarchy, we simply treat $\hat{\vz}_{l+1}$ as an additional input token and append it to $\vz_l$ before feeding into ViT. During training, we randomly replace the $\hat{\vz}_{l+1}$ with a learnable empty token with 10\% to facilitate classifier-free guidance (CFG)~\cite{ho2022classifier}. We use the same architecture for both COCO and ImageNet-1k experiments and we train on COCO for 1000 epochs and ImageNet-1k for 200 epochs if not mentioned otherwise. 

We follow the standard evaluation protocol to report the image generation quality with Fréchet inception distance (FID). For ImageNet-1k, we generate 50K images, 50 for each category, and compute the FID over the validation set using the precomputed statistic provided by \citet{dhariwal2021diffusion}. For COCO-2014, when comparing to other approaches in Table~\ref{table:compare_others_cond_coco}, we follow previous literature to generate 30k images using text prompts from the validation set and compare the statistics against the validation images.

\textbf{Hierarchical latent variables.} We build hierarchical latent variables $\{\vz_l\}_{l=2}^L$ using a pre-trained visual encoder and directly use VAE bottleneck features as our bottom levels $\vz_1$. For the ImageNet experiments, we extract visual features from MoCo-v3~\cite{chen2021empirical} (ViT-B/16), a leading self-supervised visual representation learner. For the COCO experiments, we use CLIP~\cite{radford2021learning} (ViT-B/16), a multi-modal encoder that aligns visual and textual representations. 

To ensure consistency across models with varying output dimensions, we set the top-level latent representation to a fixed dimension: $\vz_L \in \mathbb{R}^{256}$, by trimming additional feature channels via singular value decomposition (SVD).  Consequently, we construct hierarchical latent variables with shapes: $\vz_L\in \mathbb{R}^{1\times256}, \vz_{L-1}\in \mathbb{R}^{4\times128}, \vz_{L-2}\in \mathbb{R}^{16\times64}$, and so on. 

Patchification enables us to extract features at various resolutions from the encoder. However, this approach can encounter limitations when patches become too small to carry meaningful visual patterns. For cases requiring patches' spatial resolution smaller than $64\times64$, we use the feature maps from the encoder's backbone instead.  For example, with $L = 5$, $\vz_2\in \mathbb{R}^{64\times32}$ is built by reducing the feature map of the encoder, with shape $14\times14\times768$ (height $\times$ width $\times$ channel numbers) reduced to $8\times8\times32=64\times32$ through spatial pooling and channel reduction.

\begin{figure*}[t]
    \centering

    \begin{minipage}{0.05\textwidth}\raggedright
    {\small $L = 2$}
    \end{minipage}\hfill
    \begin{minipage}{0.95\textwidth}
        \centering
        \includegraphics[width=1.0\textwidth]{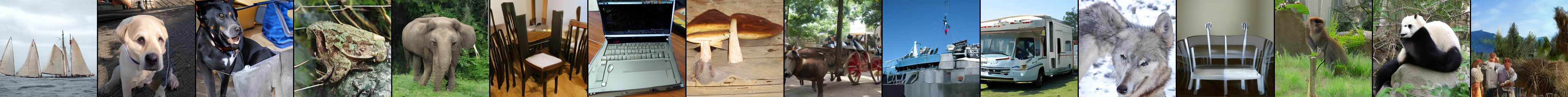}
    \end{minipage}

    \begin{minipage}{0.05\textwidth}\raggedright
    {\small $L = 3$}
    \end{minipage}\hfill
    \begin{minipage}{0.95\textwidth}
        \centering
        \includegraphics[width=1.0\textwidth]{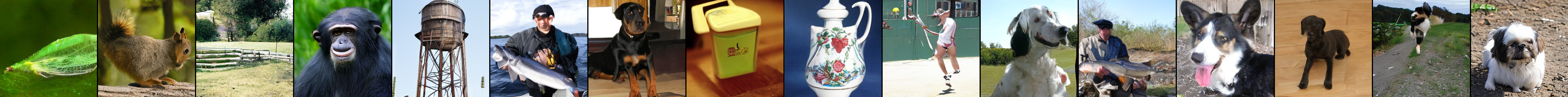}
    \end{minipage}
    
    \begin{minipage}{0.05\textwidth}\raggedright
    {\small $L = 4$}
    \end{minipage}\hfill
    \begin{minipage}{0.95\textwidth}
        \centering
        \includegraphics[width=1.0\textwidth]{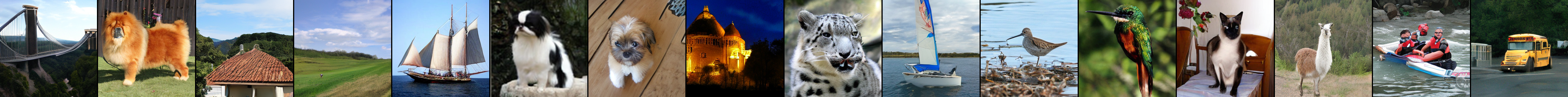}
    \end{minipage}
    
    \begin{minipage}{0.05\textwidth}\raggedright
    {\small $L = 5$}
    \end{minipage}\hfill
    \begin{minipage}{0.95\textwidth}
        \centering
        \includegraphics[width=1.0\textwidth]{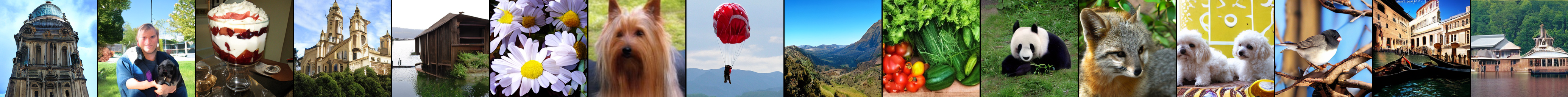}
    \end{minipage}
    \caption{\textbf{Visualization of unconditional image generation on ImageNet-1K.} We present visualizations of images generated by hierarchical diffusion models containing from $2$ to $5$ levels, demonstrating that image quality improves as the depth of the hierarchy increases.}
\end{figure*} 
\begin{table}
\centering
\begin{tabular}{l|c|c}
    Model & GFlops & FID $\downarrow$ \\
    \hline
    DiT-L/2~\cite{peebles2023scalable} & 80.0 & 23.3 \\ DiT-L/2~\cite{peebles2023scalable} + REPA~\cite{yu2024representation} & 80.0 &  15.6 \\
    \hline
    DiT-XL/2~\cite{peebles2023scalable} & 118.6 &19.5 \\ DiT-XL/2~\cite{peebles2023scalable} + REPA~\cite{yu2024representation} & 118.6 & 12.3\\
    \hline
    Baseline $L=1$~\cite{bao2023all} & 26.8 & 31.1 \\ Ours $L=5$ 
    & 34.0 & \textbf{9.9} \\ 
    \hline
    Ours $L=5$ (Unconditional) & 34.0 & \textbf{11.1}\\
\end{tabular}
\caption{\textbf{Comparison of generation quality on ImageNet-1k 256$\times$256}. When compared to other diffusion models such as DiT-L/2, XL/2, and REPA for class-conditional image generation without CFG, our models demonstrate substantial improvements while requiring fewer GFLOPs. Notably, our unconditional model outperforms their class-conditional approaches by a significant margin.
}
\label{table:repa}
\end{table}

\begin{table}
    \centering
    \small
    \begin{tabular}{c|c|c|c|c|c}
     $\gamma$ &  0 & 0.1 & 0.3 & 0.5 & $\infty$ \\
     \hline
     w/ CFG & 5.09 &  5.05 & \textbf{5.03} & 5.33 & 11.21\\
     \hline
     w/o CFG & 14.19 & 13.53 & 12.51 & 11.93 & \textbf{11.27}
\end{tabular}

    \caption{ We investigate the effects of different values of $\gamma$, which controls the level of noise added to $\vz_l$ during generation through the term $(t / T)^\gamma \sigma_l$ at diffusion steps $t$, as described in Sec.~\ref{sec:noise_feature}. Here, we present the generation results (FID) using a five-level nested diffusion model ($L=5$) on ImageNet-1K. When $\gamma = 0$, the model applies $\sigma_l$ (consistent with the noise level during training) for generation, while $\gamma = \infty$ corresponds to no noise being added to $\vz_l$ during generation.}
    \label{tab:gamma}
\end{table}

\textbf{Efficient training and parameter search for hierarchical models.} 
In our design, we allow each $\mD_{\theta_l}$ to have a unique $\sigma_l$, providing flexibility, but also adding complexity to the hyper-parameter search. To streamline this process, we use a hierarchical local search strategy: for a $L$-level model, we retain the optimal noise levels $\{\sigma_l\}_{l=3}^L$ from the $L-1$ level model and search only for the newly introduced level $\sigma_2$. An additional benefit of this approach is that we can directly reuse the parameters $\{\mD_{\theta_l}\}_{l=3}^L$ from shallower models due to the consistent model configuration, which means that we only need to train $\{\mD_{\theta_l}\}_{l=1}^2$.

\subsection{Generation with noisy hierarchical features}
\label{sec:noise_feature}
During training, we introduce noise to hierarchical features $\vz_l$ by sampling $\hat{\vz}_{l}\sim \mathcal{N}(\vz_l, \sigma{l}^2\mathbf{I})$, as outlined in Section \ref{sec:latent_variable}, to enhance information compression. By default, this noise is removed during generation to ensure consistent conditional signals. However, we observe that this approach can be detrimental to Classifier-Free Guidance (CFG), likely due to a distributional shift between training and testing. To address this, we propose a gradual decay of $\sigma_l$ across diffusion steps $t$ during generation: $\hat{\vz}_{l}^{(t)}\sim \mathcal{N}(\vz_l, (t / T)^{\gamma} \cdot \sigma_{l}^2\mathbf{I})$, where $0\leq (t / T)^{\gamma} \leq 1$ progressively reduces $\sigma_l$, and the scalar $\gamma \geq 0$ controls the decay speed. This design shares the similar spirit of \citet{sadat2023cads}, which adds noise to the ground truth image label to encourage diversity in the generation process. In contrast, our approach focuses on maintaining consistency between the training and testing phases.

Table~\ref{tab:gamma} presents the results for various $\gamma$ values in a $L=5$ nested diffusion model for unconditional generation on ImageNet-1K. Setting $\gamma = 0$ applies the same noise level $\sigma_l$ used during training, while $\gamma = \infty$ eliminates noise during generation. In experiments with CFG, adding noise to $\vz_l$ is crucial, and our proposed scheme improves generation quality compared to the baseline of using the training noise level ($\gamma = 0$). Without CFG, the best results are achieved by omitting noise from $\vz_l$. For subsequent experiments, we use $\gamma = 0.3$ in CFG scenarios and $\gamma = \infty$ in non-CFG cases.

\subsection{Benchmarking generation quality}
We present our primary results for ImageNet-1k in Table~\ref{table:big_table}, including the choices of model depth $L$, the noise level, and generation w/ or w/o CFG. 

\textbf{Improved performance with more hierarchy levels $L$.} Compared to the baseline model, our nested diffusion models demonstrate enhanced image quality as we deepen the hierarchical structure by increasing depth $ L $. Specifically, our five-level generation significantly outperforms the baseline with $ L = 1 $ in both conditional generation ($ 31.13 \rightarrow 9.87 $) and unconditional generation ($ 55.41 \rightarrow 11.05 $). Notably, our unconditional generator with $ L = 5 $ surpasses the conditional baseline generation model with $ L = 1 $, achieving scores of $ 31.13 \rightarrow 11.05 $.

We also present results using CFG, which directs the generated output toward conditional features, enhancing quality at the expense of reduced output diversity. The influence of CFG is controlled by the parameter $ w $. Our nested diffusion models produce hierarchical representations; ideally, the top level benefits from stronger CFG to enforce semantic abstraction, while lower levels should reduce the CFG influence to promote diversity in visual details. To achieve this, we adopt a straightforward approach by specifying a set of decaying CFG weights, $ \{w_i\} = [0.5, 0.4, 0.3, 0.2, 0.1] $, and selecting $ \{w_i\}_{i = 1:L} $ for our $ L $-level nested diffusion models, with higher CFG weights assigned to the top levels. We conducted a limited hyperparameter search over the choices of CFG weights and found that this strategy yields better results than using a constant CFG weight for ImageNet conditional generation. For unconditional generation, we use a constant $ w = 0.8 $, as it produces better results.

With CFG, we demonstrate that increasing the hierarchy depth $L$ further improves image quality, with our $L = 5$ model achieving FID scores of 3.97 for conditional generation and 5.03 for unconditional generation, significantly outperforming the baseline at $L = 1$ (13.75).

As illustrated in Figure \ref{fig:scale-gflops}, our hierarchical models achieve computational efficiency by constructing hierarchical features with decreased spatial dimensions, thereby reducing computational expenses at higher tiers. 
In particular, our system with $L = 5$ results in only a 27.00\% increase in the computational load measured in GFlops compared to the baseline $L = 1$, while achieving a marked decrease in FID by 68.29\%, as detailed in Appendix \ref{app:appendix_experiments}.

\textbf{Impact of $\sigma_l$.} As described in Section \ref{sec:latent_variable}, $\sigma_l$ plays a key role in enabling our design to scale effectively with more hierarchy levels. It regulates the information conveyed by the conditional latent variable $\vz_l$, ensuring that the hierarchical model does not bypass intermediate levels. We validate this design in Table~\ref{table:big_table}, where the performance gap between $\sigma_2 = 0$ and non-zero $\sigma_2$ widens as the model depth $L$ increases. This can be attributed to the fact that as $L$ grows, more feature elements are included, increasing the likelihood of the model relying primarily on the lower-level features for generation, thus neglecting higher-level features. Higher noise levels help counteract this effect: with $L=5$, setting $\sigma_2 = 1.0$ achieves FID of 11.05 and 9.87, outperforming $\sigma_2 = 0$, which yields FID of 18.04 and 19.04 for conditional and unconditional generation, respectively.

\textbf{Comparison to other methods.} We evaluate our models on class-conditional ImageNet $256\times256$ generation tasks without CFG, with results presented in Table \ref{table:repa}. We compare against DiT~\cite{peebles2023scalable} variants and REPA~\cite{yu2024representation}, which aligns diffusion representations with a pre-trained visual encoder and is trained for 400K steps. Our model with $L=5$ significantly outperforms these baselines by a substantial margin while also requiring fewer GFlops. Remarkably, our unconditional model even surpasses their conditional version.

\begin{table}[t]
\small{
\centering
\strutlongstacks{T}
\centering
\begin{subtable}{\linewidth}
\centering
\begin{tabular}{lcc}
Model & FID & Training Dataset \\
\hline
\multicolumn{2}{l}{\color{gray}\textit{\textbf{Huge Model, Extra Data}}}\\
%DALL-E-12B~\citep{ramesh2021zero} & 28.00 & DALL-E (250M) \\ 
%CogView~\citep{ding2021cogview} & 27.10 & Internal data (30M) \\
GLIDE~\citep{nichol2021glide} & 12.24 & DALL-E (250M) \\ 
DALL-E 2~\citep{ramesh2022hierarchical} & 10.39 & DALL-E (250M)\\

Imagen~\citep{saharia2022photorealistic} & 7.27 & Internal Data/LAION (860M)\\
Re-Imagen~\citep{chen2022re} & 5.25 & KNN-ImageText/COCO(50M)\\
CM3Leon-7B~\citep{yu2023scaling} & 4.88 & Internal Data(350M) \\ 
Parti-20B~\citep{yu2022scaling} & \textbf{3.22} & LAION/FIT/JFT/COCO(4.8B) \\
\hline
\multicolumn{3}{l}{\color{gray}\textit{\textbf{COCO Data Only, w/ CFG}}}\\
VQ-Diffusion~\citep{gu2022vector}&13.86&COCO(83K)\\Friro~\citep{fan2023frido}&8.97&COCO(83K)\\
U-ViT~\citep{bao2023all} & 5.42 & COCO(83K)\\
Ours $L=2$ & \textbf{4.72} & COCO(83K)\\
Ours $L=3$ & 5.92 & COCO(83K)\\
\hline
\multicolumn{3}{l}{\color{gray}\textit{\textbf{COCO Data Only,  w/o CFG}}}\\
U-ViT~\citep{bao2023all} & 14.98 & COCO(83K)\\
Ours $L=2$ & 8.15 & COCO(83K)\\
Ours $L=3$ & \textbf{6.97} & COCO(83K)\\
\end{tabular}
\end{subtable}}
\caption{\textbf{Comparison of text-to-image generation on COCO-2014.} The upper half shows larger models trained with more data and the bottom half shows the models that are only trained on training split of COCO. When trained only on COCO, 
our models (with $\sigma_2 = 0.5$) outperform all the compared methods. It is worth noting that we're better than most of the larger models, shown on the top half.}
\label{table:compare_others_cond_coco}
\end{table}
\textbf{Experiments on COCO.} In addition to ImageNet-1K, we also evaluate our model on the COCO-2014 dataset to assess performance on complex scenes. Using visual features from the CLIP ViT-B/16 model and fixed noise levels $\{\sigma_l\} = 0.5$, Table~\ref{table:compare_others_cond_coco} reports results, comparing our approach with large models trained on additional data sources. Without CFG, our $L=3$ system achieves state-of-the-art performance, surpassing most larger models. For our models, we set the CFG weight to 1, following the default in \citet{bao2023all}. When generating with CFG, we find that $L=2$ delivers the best performance, even outperforming the 7-billion-parameter model, CM3Leon. The $L=3$ model performs worse than $L=2$, likely due to a suboptimal CFG weight.

\subsection{Generation with different visual encoders.}
In Section ~\ref{sec:semantic_neighbours}, we discuss the importance of preserving neighbor structures for the target space of diffusion models. We quantitatively validate this claim in  Table~\ref{table:cond_visual_sources} by presenting results from a 3-level nested diffusion model ($L=3$) with fixed $\{\sigma_l\} = 0.5$, applied to text-to-image generation on the COCO dataset. We construct $\vz_l$ using various visual encoders, including MAE, MoCo-v3, CLIP, and DINO. For a fair comparison, we use the same encoder architecture (ViT-B with a patch size of 16) and ensure that all $\vz_l$ representations have the same feature dimension. Our results show that image generation quality consistently improves with better visual representations, as measured by KNN accuracy on the ImageNet-1k dataset with $K=20$.

\begin{table}[t]
\centering
\strutlongstacks{T}
\centering
\begin{subtable}[t]{0.45\textwidth}
\centering
\begin{tabular}{l|c|c|c}
\multirow{2}{*}{Features} & \multirow{2}{*}{FID$\downarrow$} & \multicolumn{2}{c}{KNN Acc. $\uparrow$}\\
& &   Top1 & Top5 \\ 
\hline
None & 14.98   & - & - \\
MAE~\cite{he2022masked} &  10.96  & 27.44 &45.33 \\
MoCo-v3~\cite{chen2021empirical} & 10.59  & 66.57 & 83.09\\
CLIP~\cite{radford2021learning} &  \textbf{6.97}  & 73.35 & \textbf{91.12}\\
DINO~\cite{caron2021emerging} &  \textbf{6.78}  & \textbf{75.86} & \textbf{91.17} \\
\end{tabular}
\end{subtable}\hfil
\caption{\textbf{Results on COCO text-to-image generation with different visual representations.} We compare the generation quality of a 3-level nested diffusion model, where $L=3$ and $\{\sigma_l\} = 0.5$, using various visual encoders to construct $\vz_l$. We report the results \textit{without} CFG. Additionally, we report the accuracy of a KNN classifier with $K=20$ on ImageNet-1K to quantify feature quality. Our results indicate that better feature quality improves generation results.
}
\label{table:cond_visual_sources}
\end{table}

\section{Conclusion}
We introduce nested diffusion models, a novel hierarchical generative framework utilizing a succession of diffusion models to generate images starting from low-dimensional semantic feature embeddings and proceeding to detailed image refinements. Unlike conventional single-level latent models and hierarchical models that use low-level feature pyramids, each level in our model is conditional on higher hierarchical semantic features. This distinctive design improves image structure preservation and maintains global consistency, enhancing generation quality with minimal extra computational expense. Furthermore, we showcase the scalability of our method through a deeper unconditional system, which significantly surpasses the performance of the conditional baseline.

\section{Acknowledgements}
RJ and RW gratefully acknowledge the support
of the DOE grant DE-SC002223 and NSF grant DMS-2023109. 

{
    \small
    \bibliographystyle{ieeenat_fullname}
    \bibliography{main}
}

\newpage
\appendix
\newpage
\section{Appendix}
\label{app:appendix_experiments}
\begin{figure*}[h!]
    \centering

    \begin{minipage}{0.95\textwidth}
        \centering
        \includegraphics[width=1.0\textwidth]{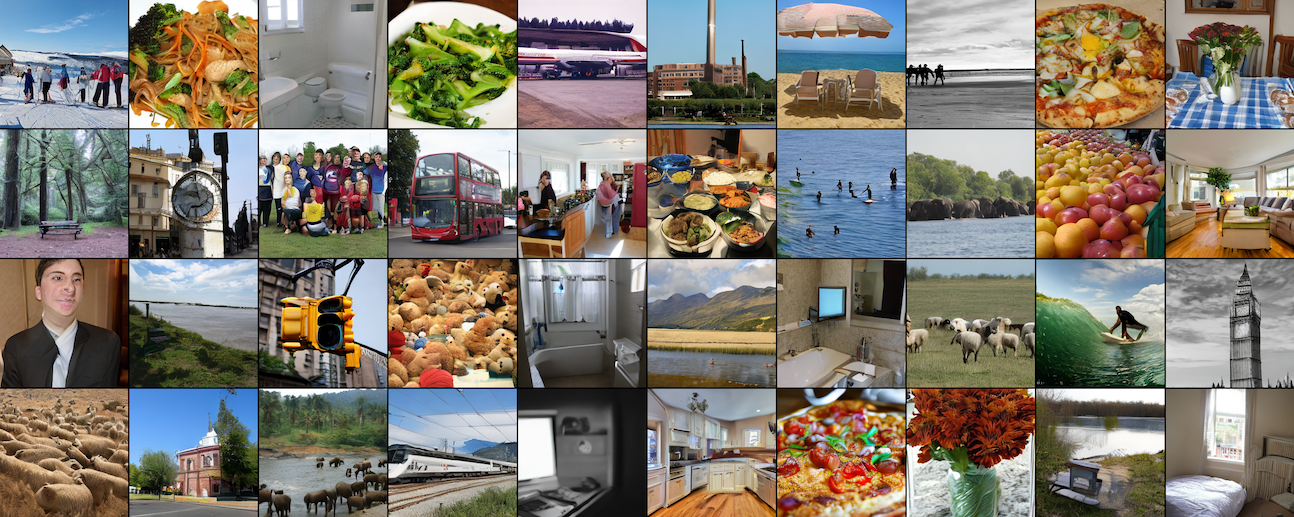}
        (a) $L=2$
    \end{minipage}

    \begin{minipage}{0.95\textwidth}
        \centering
        \includegraphics[width=1.0\textwidth]{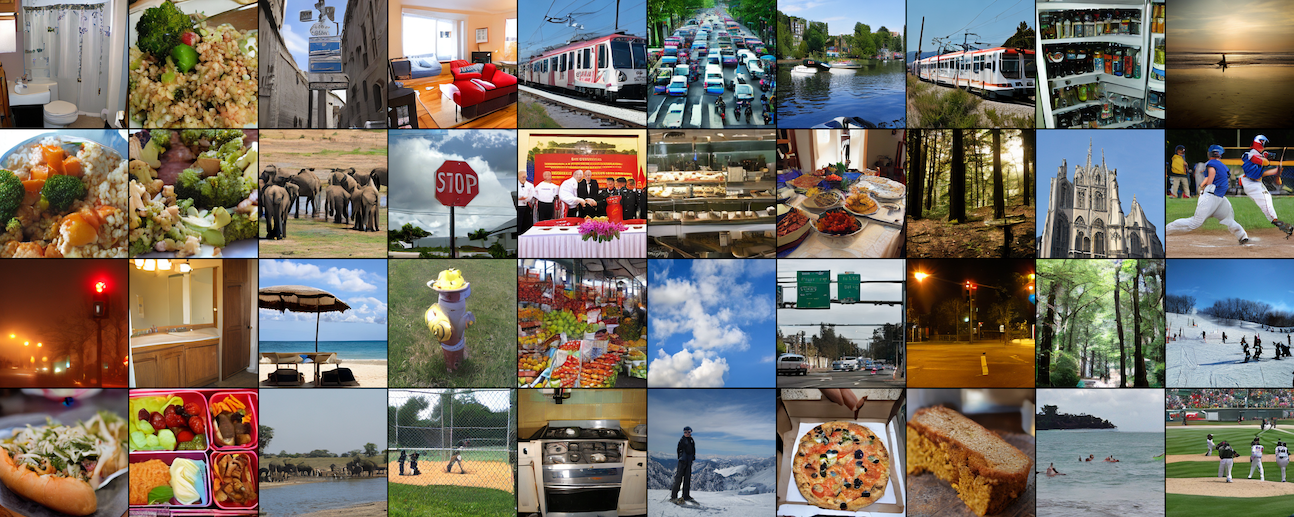}
        (b) $L=3$
    \end{minipage}
    
    \caption{\textbf{Visualization of text-to-image generation on COCO2014.} We present visualizations of images generated by hierarchical diffusion models of $2$ and $3$ levels.}
    \label{fig:appendix_coco}
\end{figure*}
\subsection{Further comparison}
\begin{table}[h]
\footnotesize
\centering
\begin{tabular}{c|c|c|c|c|c|c}
     $L$ &  1 & 2 & 3 & 4 & $5$ & $1^*$\\
     \hline
     GFlops & 26.82 & 27.13 & 28.17 & 29.61 & 33.98 & 35.11 \\
     \hline
     Param(M) & 104 & 208 & 313 & 417 & 523 & 135\\
     \hline
     FID $\downarrow$ & 31.13 & 16.52 & 15.50 & 13.87 & \textbf{9.87} & 19.74 \\
\end{tabular}
\caption{\textbf{Comparison of multiple model configurations over model depth $L$.} Unlike the baseline diffusion model ($L = 1^*$) whose computational complexity (GFlops) grows linearly with model parameters, our efficient hierarchical design only yields minimal GFlops growth with deeper models but achieves much better image quality than the baseline model with the same GFlops.}
\label{table:complexity}
\end{table}

Table~\ref{table:complexity} presents a comparison of model parameters and computational complexity for models with varying depths $L$. In contrast to the baseline model $1^*$, whose computational complexity scales linearly with the number of parameters, our efficient hierarchical design incurs only a modest computational overhead for deeper models. Under a comparable computational budget, our model with $L=5$ demonstrates significantly better performance than $1^*$.

\subsection{Derivation of formulas}
\textbf{Derivation for $\mathcal{L}_{\rm ELBO}$} (Eqn.\ref{eqn:nested_diffusion_elbo}).
Let $\vx=\vz_1$ be the observed data and $\vz_2, \vz_3, \dots, \vz_L$ be the latent variables with $\vz_{>l} := \{ \vz_m\}_{m=l+1}^L$.
We assume the joint distribution of data and latent variables can be modeled as follows:
\begin{equation}
    p_\theta(\vx, \vz_{>1}) = 
    p_\theta(\vx |  \vz_{>1})
    \prod_{l=2}^{L-1} p_\theta(\vz_l | \vz_{>l}) p_\theta(\vz_L),
    \label{eqn:joint_dist}
\end{equation}
with the corresponding posterior written as:
\begin{eqnarray}
   q(\vz_{>1}|\vx) = q(\vz_L|\vx)\prod_{l=2}^{L-1} q(\vz_l | \vz_{>l}, \vx).
   \label{eqn:posterior}
\end{eqnarray}
For the derivation of the ELBO, we proceed in a similar way as \citet{vahdat2020nvae, pervez2020variance, takida2023hq} by relying on Jensen's equality:
\begin{align}
    \log p_{\theta}(\vx) 
    &= \log \int p_\theta(\vx, \vz_{>1}) {\rm d} \vz_{>1} \\
   &= \log \int q(\vz_{>1}|\vx) \frac{p_\theta(\vx, \vz_{>1})}{q(\vz_{>1}|\vx)} {\rm d} \vz_{>1} \\ 
   & \geq \mathbb{E}_{q(\vz_{>1}|\vx)} \log \frac{p_\theta(\vx, \vz_{>1})}{q(\vz_{>1}|\vx)} \\
   & \equiv {\rm ELBO}.
\end{align}
By plugging in Eqn.\ref{eqn:joint_dist} and Eqn.\ref{eqn:posterior}:
\begin{align}
   {\rm ELBO}
   =& \mathbb{E}_{q(\vz_{>1}|\vx)}  \bigg[ \log p_\theta(\vx | \vz_{>1}) 
   \\
   &+ \sum_{l=2}^{L-1} \log \frac{p_\theta(\vz_l | \vz_{>l})}{q(\vz_l | \vz_{>l}, \vx)} + \log \frac{p_\theta(\vz_L)}{q(\vz_L | \vx)} \bigg] \nonumber \\
   \label{eqn:derived_elbo}
   =& \mathbb{E}_{q(\vz_{>1}|\vx)} \log p_\theta(\vx | \vz_{>1})  \\ 
   &-\sum_{l=2}^{L-1} 
   \mathbb{E}_{q(\vz_{>l}| \vx)} 
   D_{\rm KL}\left( q(\vz_l | \vz_{>l}, \vx) | p_\theta(\vz_l | \vz_{>l})\right) 
   \nonumber 
   \\
   &-D_{\rm KL} \left(q(\vz_L | \vx) | p_\theta(\vx_L) \right).\nonumber
\end{align}
To incorporate diffusion models to parameterize $p_\theta(\vz_l | \vz_{>l})$, we further decompose the KL divergence for each level $l$.
Since we utilize a pre-trained encoder that computes each latent variable $\vz_l$ directly from the observed data $\vx$ (see Section~\ref{sec:latent_variable}), we can simplify the conditional posterior distribution by removing the dependence on $\vz_{>l}$:
\begin{align}
    &-D_{\rm KL}\left( q(\vz_l | \vz_{>l}, \vx) | p_\theta(\vz_l | \vz_{>l})\right) \\
   =& \int q(\vz_l | \vx)
    \bigg[
   \log p_\theta(\vz_l | \vz_{>l})
   -
   \log q(\vz_l | \vx)
   \bigg] {\rm d} \vz_l.
\end{align}
As the posterior $q$ has no learnable parameters $\theta$, then maximizing the negative KL divergence equals maximizing
\begin{eqnarray}
    \int q(\vz_l | \vx)
   \log p_\theta(\vz_l | \vz_{>l})
    {\rm d} \vz_l.
    \label{eqn:simplified_data}
\end{eqnarray}
Now assume that the latent variable $\vz_l$ is modeled through a diffusion process:
\begin{eqnarray}
    p_\theta(\vz_l | \vz_{>l}) = \int p_\theta(\vz_l^{(0:T)} | \vz_{>l}) \rm{d} \vz_l^{(1:T)},
\end{eqnarray}
where $\vz_l^{(0)} = \vz_l$, and $\vz_l^{(t)}$ denotes the noise latent variable at time step $t, \forall t \in \{0, 1, \dots, T\}$.
Then maximizing the likelihood in Eqn.\ref{eqn:simplified_data} amounts to maximizing
\begin{align}
    & \int q(\vz^{(0)}_l | \vx) \log p_\theta(\vz^{(0)}_l | \vz_{>l}) {\rm d} \vz^{(0)}_l
    \nonumber \\
    =& \int {\rm d} \vz^{(0)}_l q(\vz^{(0)}_l | \vx)   \\
    &\log \Bigg[ {\rm d} \vz_l^{(1:T)} 
    \frac{p_\theta(\vz^{(0:T)}_l | \vz_{>l})}
    {q(\vz_l^{(1:T)} | \vz^{(0)}, \vx)} 
    q(\vz_l^{(1:T)} | \vz^{(0)},\vx) \Bigg] \nonumber \\
    % = & \int {\rm d} \vz^{(0)}_l q(\vz^{(0)}_l | \vx)   \\
    % & \log \Bigg[ {\rm d} \vz_l^{(1:T)}
    % p(\vz^{(T)} | \vz_{>l})
    % q(\vz_l^{(1:T)} | \vz^{(0)},\vx)
    % \prod_{t=1}^T \frac{p_\theta(\vz^{(t-1)} | \vz^{(t)}, \vz_{>l})}{q(\vz^{(t)} | \vz^{(t-1)}, \vx)} \Bigg] \nonumber \\
    \geq & \int {\rm d}  \vz^{(0:T)}_l q(\vz^{(0:T)}_l | \vx) \\
    & \log \Bigg[p_\theta(\vz_l^{(T)} |\vz_{>l}) \prod_{t=1}^T \frac{p_\theta(\vz_l^{(t-1)} | \vz_t^{(t)}, \vz_{>l})}{q(\vz_t^{(t)} | \vz_t^{(t-1)}, \vx)} \Bigg] \nonumber \\
    \equiv & -\mathcal{L}_l.
\end{align}
Following the derivation in Sohl-Dickstein et al.\footnote{[82] Sohl-Dickstein et al, Deep unsupervised learning using nonequilibrium thermodynamics. ICML, 2015.}, the loss at each level $l$ can be further reduced as
\begin{eqnarray}
    \mathcal{L}_l \leq &
    \sum_t
    \int {\rm d} \vz_l^{(0)}\vz_l^{(t)} q(\vz_l^{(0)},\vz_l^{(t)}) \\
    & D_{\rm KL}
    \left(
    q(\vz_l^{(t-1)} | \vz_l^{(t)}, \vz_l, \vx) \| p_\theta(\vz_l^{(t-1)} | \vz_l^{(t)}, \vz_{>l})
    \right) \nonumber.
\end{eqnarray}
By plugging this reduced form of the loss at each level into Eqn.\ref{eqn:derived_elbo}, we arrive at Eqn.\ref{eqn:nested_diffusion_elbo}.

\subsection{Qualitative evaluation}
\begin{figure*}[t]
    \centering

    \begin{minipage}{0.95\textwidth}
        \centering
        \includegraphics[width=1.0\textwidth]{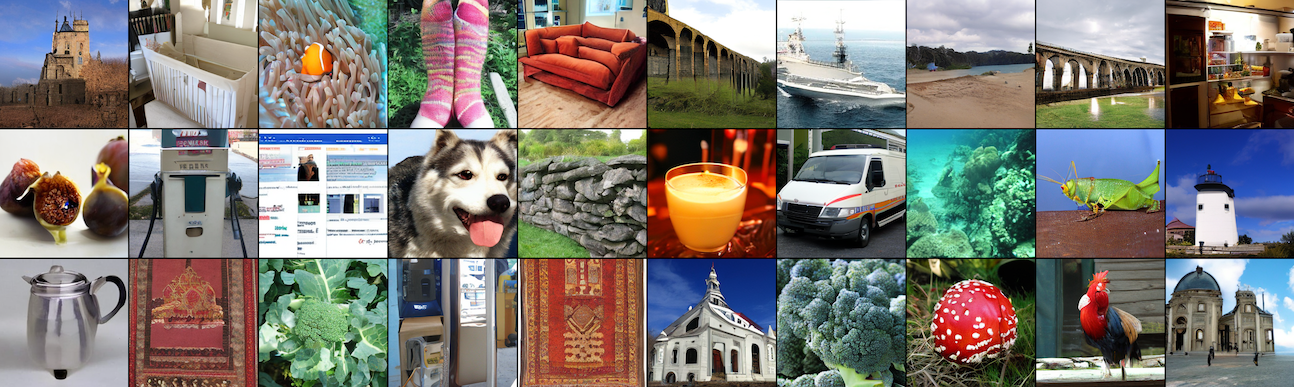}
        (a) $L=2$
    \end{minipage}

    \begin{minipage}{0.95\textwidth}
        \centering
        \includegraphics[width=1.0\textwidth]{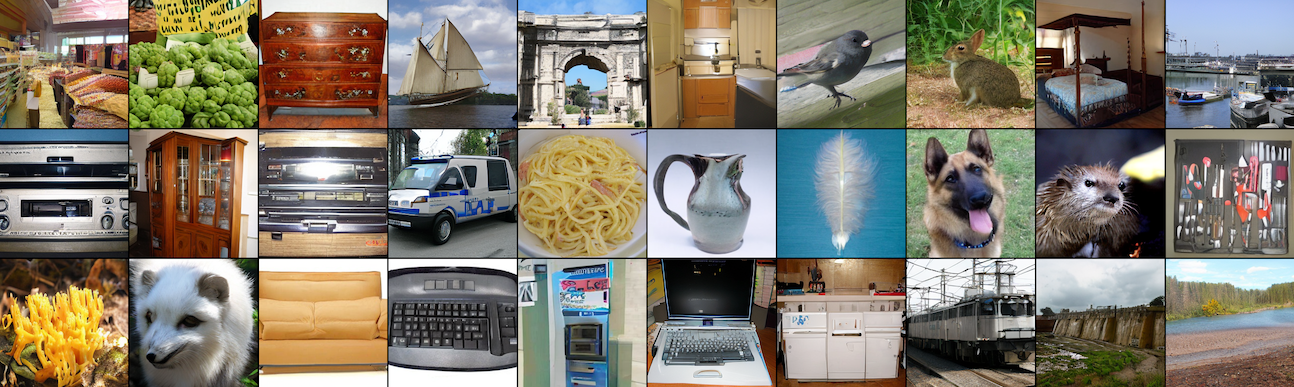}
        (b) $L=3$
    \end{minipage}
    
    \begin{minipage}{0.95\textwidth}
        \centering
        \includegraphics[width=1.0\textwidth]{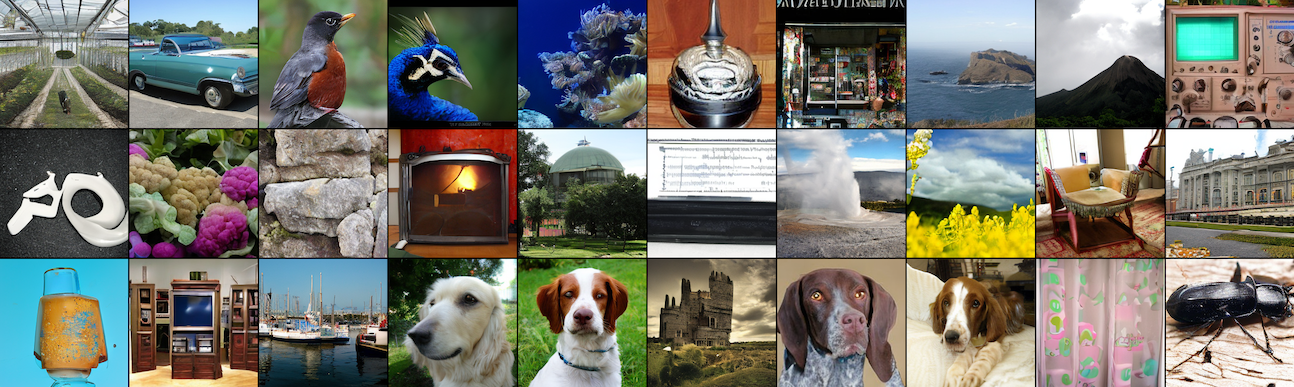}
        (c) $L=4$
    \end{minipage}
    
    \begin{minipage}{0.95\textwidth}
        \centering
        \includegraphics[width=1.0\textwidth]{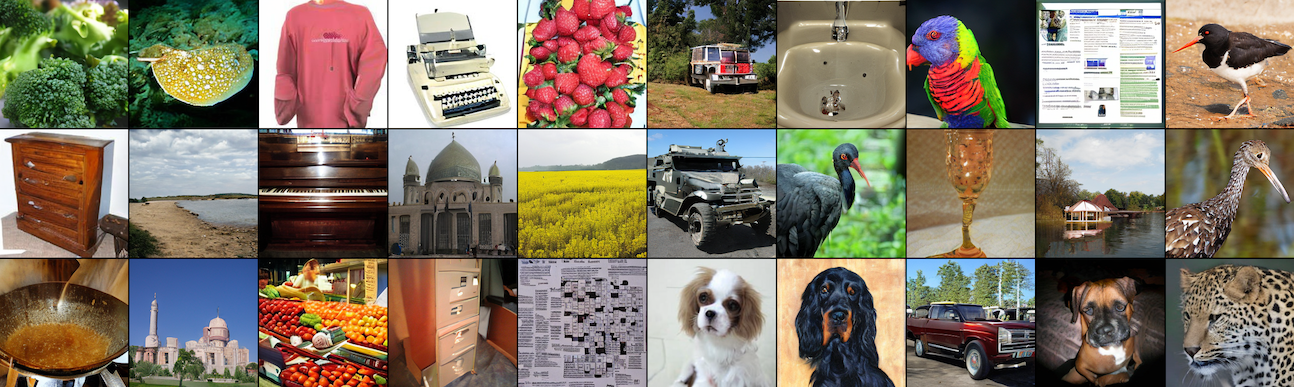}
        (d) $ L=5$
    \end{minipage}
    \caption{\textbf{Visualization of conditional image generation on ImageNet-1K.} We present visualizations of images generated by hierarchical diffusion models containing from $2$ to $5$ levels.}
    \label{fig:appendix_1k_cond}
\end{figure*}

\begin{figure*}[t]
    \centering

    \begin{minipage}{0.95\textwidth}
        \centering
        \includegraphics[width=1.0\textwidth]{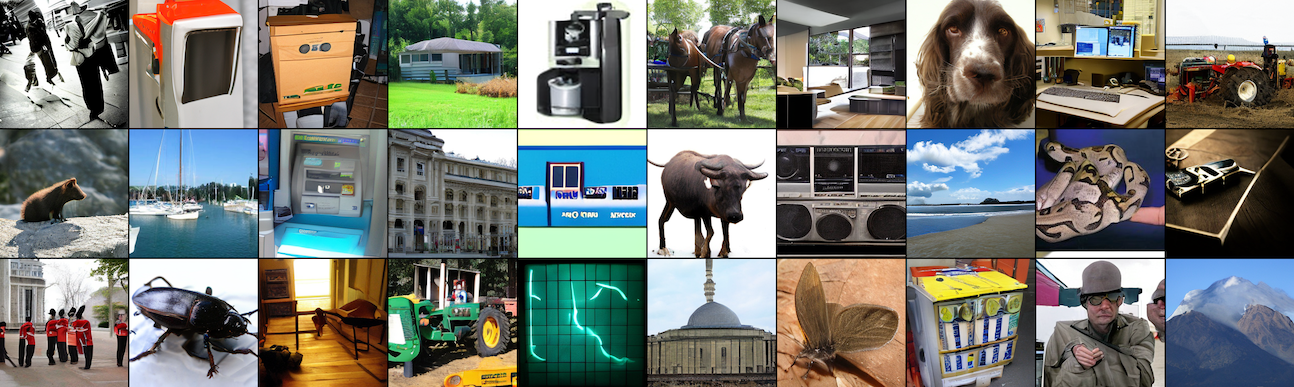}
        (a) $L=2$
    \end{minipage}

    \begin{minipage}{0.95\textwidth}
        \centering
        \includegraphics[width=1.0\textwidth]{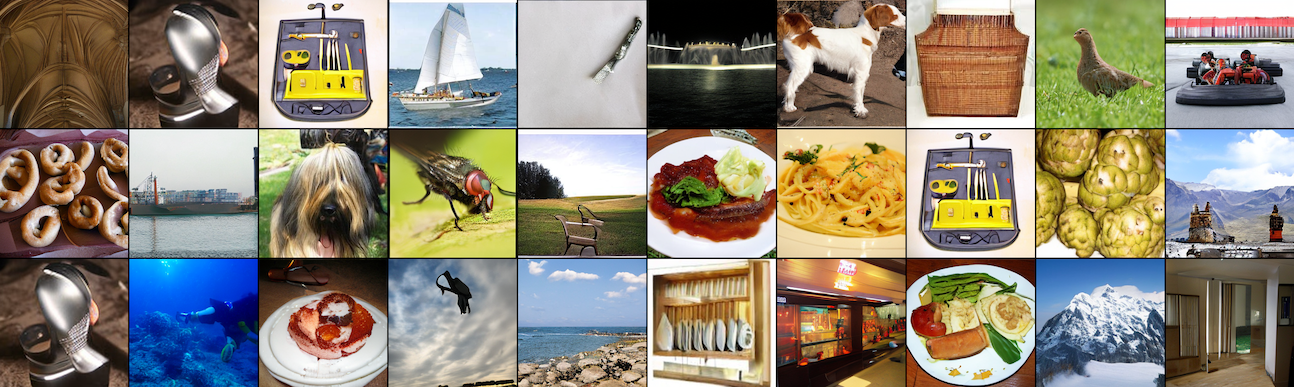}
        (b) $L=3$
    \end{minipage}
    
    \begin{minipage}{0.95\textwidth}
        \centering
        \includegraphics[width=1.0\textwidth]{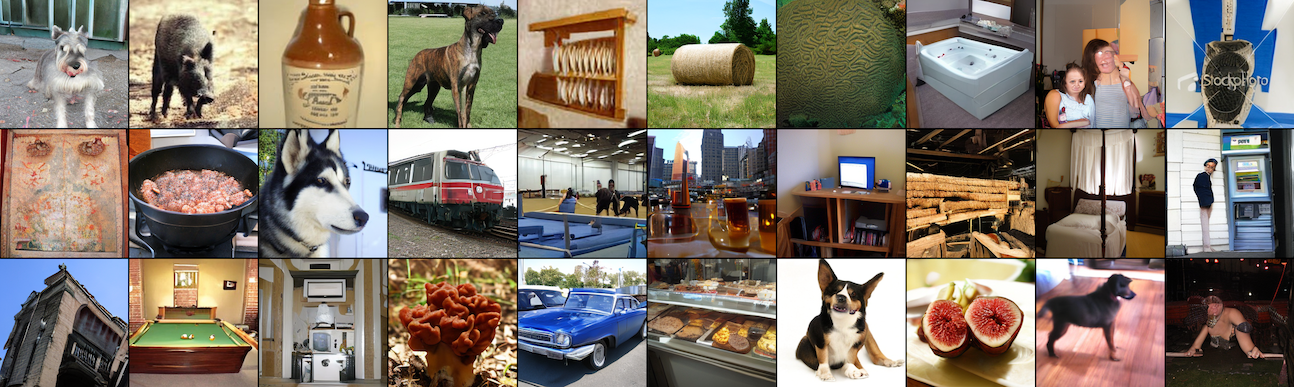}
        (c) $L=4$
    \end{minipage}
    
    \begin{minipage}{0.95\textwidth}
        \centering
        \includegraphics[width=1.0\textwidth]{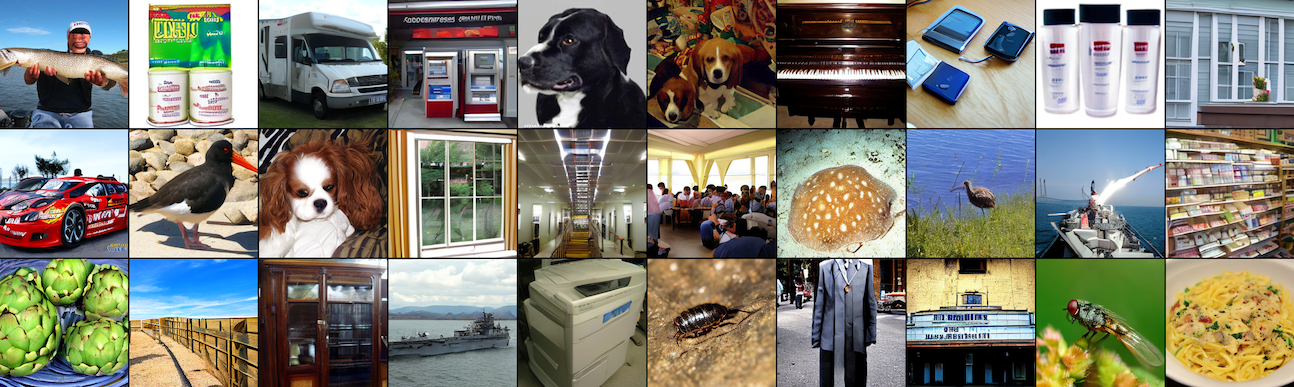}
        (d) $ L=5$
    \end{minipage}
    \caption{\textbf{Visualization of unconditional image generation on ImageNet-1K.} More visualizations of images generated by hierarchical diffusion models containing from $2$ to $5$ levels.}
    \label{fig:appendix_1k_uncond}
\end{figure*}

Additional visualizations of images generated by our model at different depths are provided: 
Fig.~\ref{fig:appendix_coco} illustrates text-to-image generation on the COCO2014 dataset, Fig.\ref{fig:appendix_1k_cond} displays conditional generation on ImageNet-1k, and  Fig.\ref{fig:appendix_1k_uncond} displays unconditional generation on ImageNet-1k.


\end{document}